  \RequirePackage{fix-cm}
  \documentclass[twocolumn]{svjour3}          
  \smartqed  

  \usepackage{soul}
  \usepackage{url}
  \usepackage[utf8]{inputenc}
  \usepackage[small]{caption}
  \usepackage{booktabs}
  \usepackage{natbib}
  
  \usepackage[colorlinks=true]{hyperref}
  \usepackage{amsmath}
  \usepackage{amssymb}
  \usepackage{xspace}
  \usepackage{tablefootnote}
  \usepackage{graphicx}
  \usepackage[ruled,vlined]{algorithm2e}
  \usepackage[table,xcdraw]{xcolor}
  \usepackage{multirow}
  \usepackage{subcaption}
  \usepackage{float}
  \usepackage[misc]{ifsym}
  \hypersetup{colorlinks, breaklinks, citecolor=[rgb]{0,0.45,0.75}, linkcolor=[rgb]{1,0,0}}
  \graphicspath{{./figures/}}
\begin{document}

\title{Weakly-Supervised Semantic Segmentation with Visual Words Learning and Hybrid Pooling}

\author{Lixiang Ru $^{1,2}$ \and Bo Du $^{1,2}$ \and Yibing Zhan $^3$ \and Chen Wu $^4$}

\institute{
  \Letter \ Bo Du (\textit{Corresponding Author.}) \at \email{dubo@whu.edu.cn} \and
  $^1$ Institute of Artificial Intelligence, School of Computer Science, Wuhan University, Wuhan, China\\
  $^2$ National Engineering Research Center for Multimedia Software, Hubei Key Laboratory of Multimedia and Network Communication Engineering, Wuhan University, Wuhan, China \\
  $^2$ JD Explore Academy, JD.com, Beijing, China\\
  $^3$ LIESMARS, Wuhan University, Wuhan, China\\
}

\date{Received: date / Accepted: date}

\maketitle

\makeatletter
\DeclareRobustCommand\onedot{\futurelet\@let@token\@onedot}
\def\@onedot{\ifx\@let@token.\else.\null\fi\xspace}

\def\eg{\emph{e.g}\onedot} \def\Eg{\emph{E.g}\onedot}
\def\ie{\emph{i.e}\onedot} \def\Ie{\emph{I.e}\onedot}
\def\cf{\emph{c.f}\onedot} \def\Cf{\emph{C.f}\onedot}
\def\etc{\emph{etc}\onedot} \def\vs{\emph{vs}\onedot}
\def\wrt{w.r.t\onedot} \def\dof{d.o.f\onedot}
\def\etal{\emph{et al}\onedot}
\makeatother

\begin{abstract}
  Weakly-Supervised Semantic Segmentation (WSSS) methods with image-level labels generally train a classification network to generate the Class Activation Maps (CAMs) as the initial coarse segmentation labels. However, current WSSS methods still perform far from satisfactorily because their adopted CAMs 1) typically focus on partial discriminative object regions and 2) usually contain useless background regions. These two problems are attributed to the sole image-level supervision and aggregation of global information when training the classification networks.
  In this work, we propose the visual words learning module and hybrid pooling approach, and incorporate them in the classification network to mitigate the above problems. In the visual words learning module, we counter the first problem by enforcing the classification network to learn fine-grained visual word labels so that more object extents could be discovered. Specifically, the visual words are learned with a codebook, which could be updated via two proposed strategies, \ie learning-based strategy and memory-bank strategy. The second drawback of CAMs is alleviated with the proposed hybrid pooling, which incorporates the global average and local discriminative information to simultaneously ensure object completeness and reduce background regions.
  We evaluated our methods on PASCAL VOC 2012 and MS COCO 2014 datasets. Without any extra saliency prior, our method achieved 70.6\% and 70.7\% mIoU on the $val$ and $test$ set of PASCAL VOC dataset, respectively, and 36.2\% mIoU on the $val$ set of MS COCO dataset, which significantly surpassed the performance of state-of-the-art WSSS methods.
\end{abstract}
\keywords{Weakly-Supervised Semantic Segmentation \and Visual Words Learning \and Hybrid Pooling \and Semantic Segmentation}

\section{Introduction}
\par {Semantic} segmentation, aiming at assigning a specific label for each pixel in an image, is a fundamental and hot topic in computer vision. With the rapid development of deep learning, semantic segmentation based on deep neural networks has dominated the past decades \citep{long2015fully,chen2014semantic,badrinarayanan2017segnet,chen2017deeplab}. However, the data-hungry nature of deep models determines that to obtain a segmentation model with fancy performance, a large number of images with well-annotated pixel-level labels are indispensable. Unfortunately, pixel-level labels are usually very costly in both time and money. The empirical statistics in \citep{lin2020block} show that annotating the pixel-level label of an image in the PASCAL VOC dataset \citep{everingham2010pascal} needs about 4 minutes on average, meanwhile annotating the Cityscapes dataset \citep{cordts2016cityscapes} takes an even longer time, about 90 minutes per image.

\par To address the above problem, many researchers have dedicated to devising image segmentation models with weaker and cheaper labels, such as image-level labels \citep{papandreou2015weakly,pinheiro2015image,ahn2018learning,lee2021anti}. Prevailing WSSS methods with image-level labels usually adopt a multi-step framework. Specifically, these WSSS methods firstly train classification networks with only image-level labels and use the trained classification networks to generate initial coarse pixel-level labels by class activation mapping \citep{zhou2016learning}. Then, the coarse pixel-level labels will be further refined by methods like dense CRF \citep{krahenbuhl2011efficient} and other pixel affinity-based approaches \citep{ahn2018learning,ahn2019weakly} to obtain the refined pseudo labels. Finally, the refined pseudo labels are used to train a regular semantic segmentation model to predict pixel-level labels of test images.

\begin{figure}[tbp]
  \centering
  \includegraphics[width=0.46\textwidth]{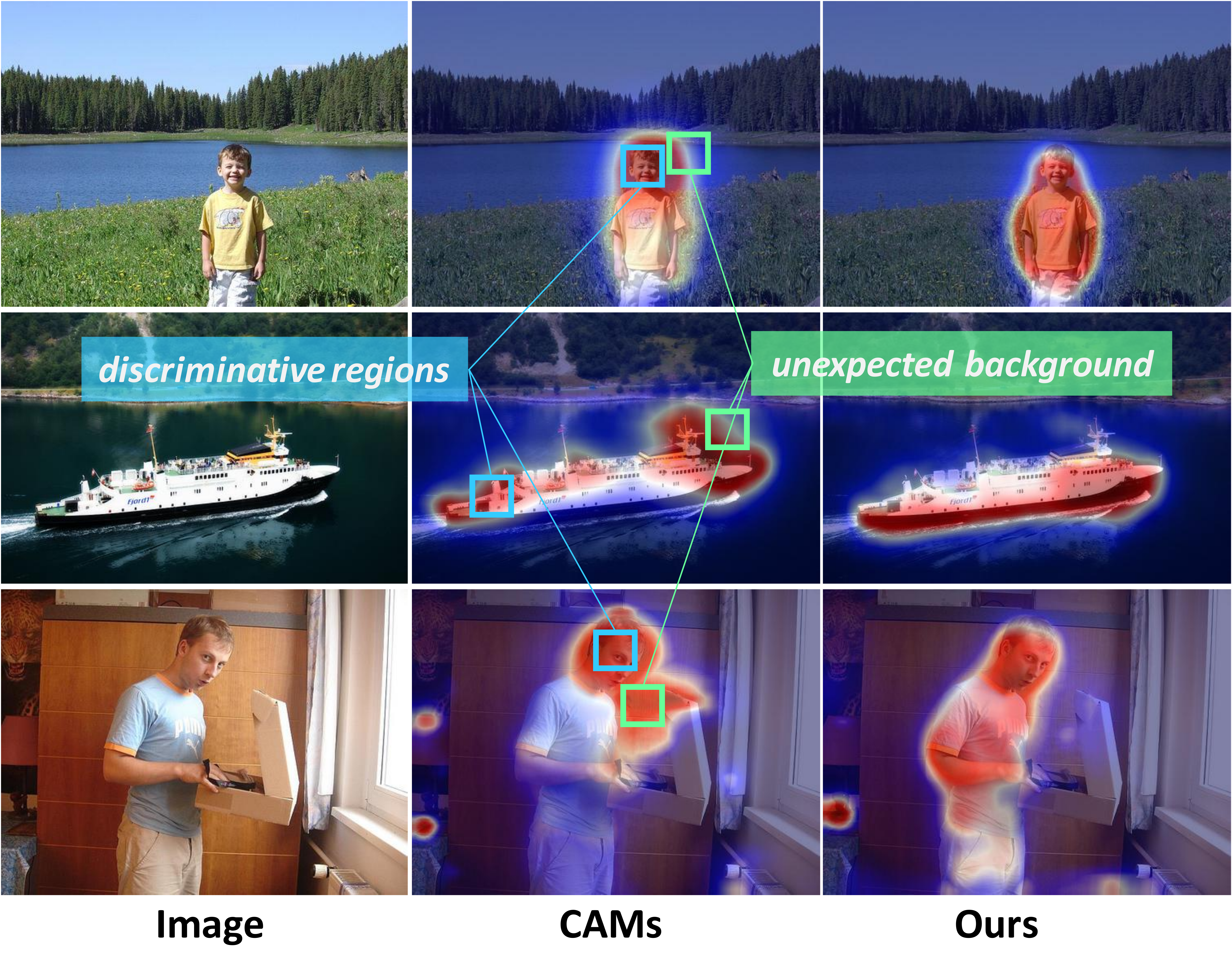}
  \caption{Illustration of the drawbacks of CAMs. Typically, CAMs only discover partial discriminative object regions and adjacent background regions. We argue that these drawbacks are attributed to the sole image-level supervision and aggregation of global information. To mitigate them, in this work, we proposed the visual words learning and hybrid pooling module.}
  \label{fig_cam}
\end{figure}

\par Prior works have demonstrated that the first step, \ie generating initial coarse labels, is crucial to the training of segmentation models and the final segmentation performance \citep{wang2020self,chang2020weakly,lee2021anti}. As aforementioned, most methods train classification networks to produce Class Activation Maps (CAMs) \citep{zhou2016learning} as the initial coarse labels.
However, as illustrated in Fig.~\ref{fig_cam}, there are two typical drawbacks of previous CAMs. Firstly, CAMs usually only discover partial discriminative regions of visual objects. The reason is that CAMs are derived from classification networks, whose purpose is to differentiate different semantic categories. Therefore, to attain discriminability, the classification network will shift attention to the most discriminative regions of visual objects instead of the integral object. Secondly, the activated regions of CAMs often include some undesired background. This is attributed to that classification networks commonly use global average pooling (GAP) \citep{lin2013network} for feature aggregation, which averages information from both foreground objects and background, thus overestimating the size of objects \citep{zhou2016learning}.

\begin{figure}[tbp]
  \centering
  \includegraphics[width=0.38\textwidth]{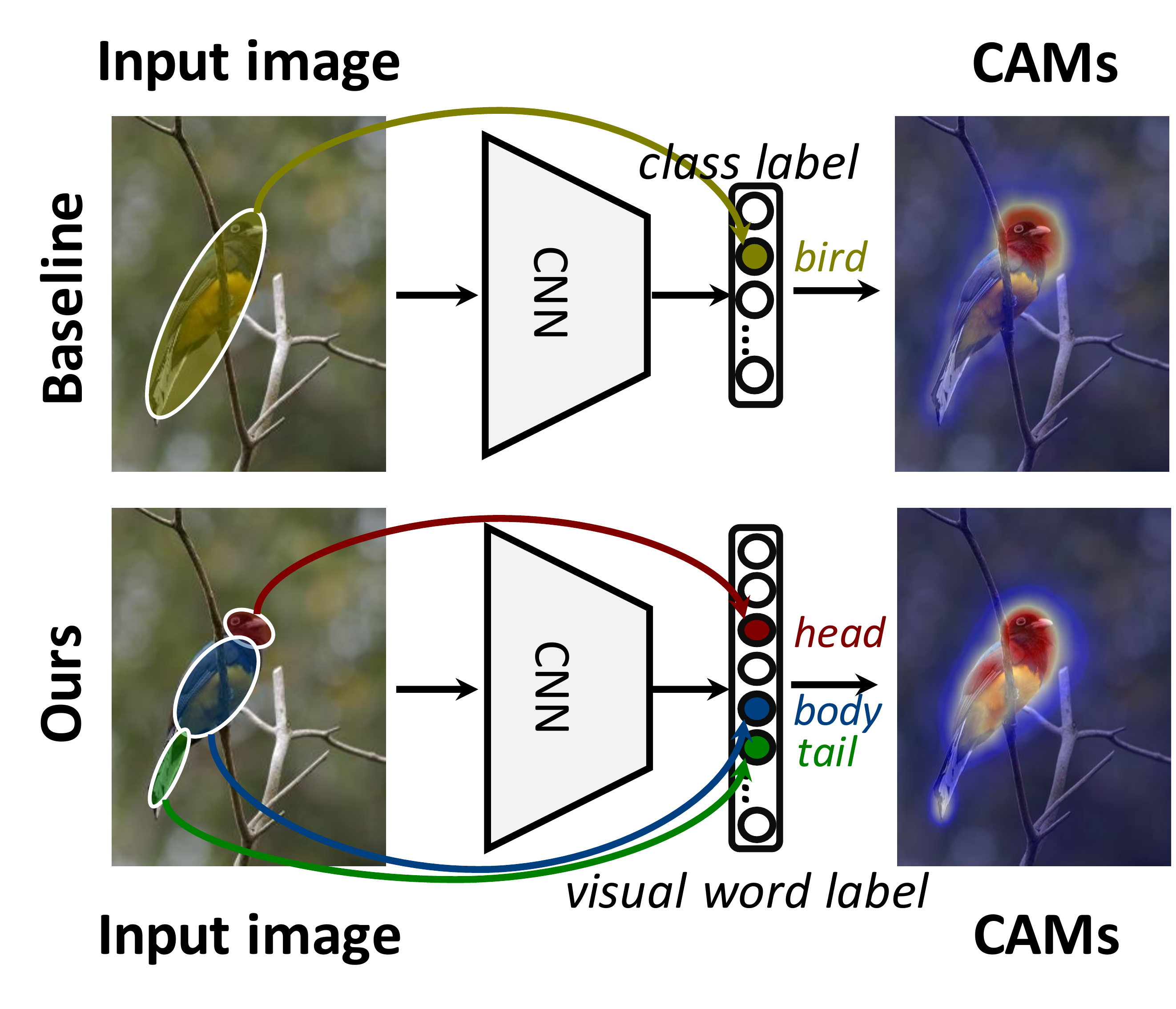}
  \caption{Illustration of the motivation of our visual words learning module. }
  \label{fig_visual_words}
\end{figure}

\par To tackle the first problem, as illustrated in Fig.~\ref{fig_visual_words}, we argue that if the network could be supervised by more fine-grained labels, more object regions will be activated to provide sufficient information for differentiating different classes. Therefore, in this work, we propose the Visual Words Learning (VWL) module for WSSS task with image-level labels. The VWL module generates the visual word labels by using a codebook to encode the feature maps extracted by the CNN backbone. In the training process of the classification network, the network will be forced to jointly learn the image-level labels and visual word labels so that more object regions could be activated. To learn an effective codebook, based on the definition and solution of Bag of Visual Word models (BoVW) \citep{arandjelovic2017netvlad,passalis2017learning}, we devised two strategies for updating the codebook, \ie learning-based strategy and memory-bank strategy. For the learning-based strategy, the codebook is set as a learnable parameter. By enforcing the encoded visual word features to learn image-level labels, the codebook could learn the latent visual word representations. In practice, we also notice that the learned representations in the codebook are often redundant, which affects the network training and the quality of CAMs. We tackle this problem by regularizing the codebook with DeCov loss \citep{cogswell2015reducing}, which reduces the redundancy of a matrix by minimizing its off-diagonal co-variance values. For the memory-bank strategy, we follow the classic BoVW models \citep{liu2019bow,gidaris2020learning}, which take the clustering centroids of features as the codebook. Specifically, we decompose the clustering on the whole training set to each mini-batch iteration and leverage memory-bank strategy \citep{wu2018unsupervised,zhuang2019local} to gradually update the codebook. Our experimental results show that, after sufficient updates, the learning-based and memory-bank strategy could both learn codebooks with effective representations of visual words and achieve analogous performance.

\par To alleviate the second drawback, inspired by global max-pooling (GMP), which takes the maximum value in each feature map as outputs and tends to underestimate the object sizes \citep{kolesnikov2016seed}, we proposed a simple yet empirically effective pooling approach, named Hybrid Pooling (HP). Our major motivation of HP is to aggregate the local maximum values so that less background information is involved. In specific, the feature maps are partitioned into multiple bins from coarse to fine levels. For bins in the same level, we pool them separately via max pooling and average the aggregated features so that only local maximums are involved. The features from different levels and the output feature of GAP (to ensure the object completeness) are then averaged as the final outputs. On this account, more discriminative object extents and fewer background regions are preserved in feature maps, which could improve the accuracy of the generated CAMs.

\par We conducted extensive experiments to verify the effectiveness of our proposed methods. The experimental results showed that our method significantly improved the quality of generated CAMs. We refined the generated CAMs with IRNet \citep{ahn2019weakly} and trained a DeepLabV2 segmentation network \citep{chen2017deeplab} with the refined pseudo labels. Semantic segmentation results on two datasets, PASCAL VOC 2012 \citep{everingham2010pascal} and MS COCO 2014 \citep{lin2014microsoft}, showed that our proposed method could outperform the state-of-the-art methods.
\par Overall, our contributions in this work are summarized as follows.
\begin{itemize}
  \item We propose the Visual Words Learning (VWL) module. By jointly learning the visual word labels and image-level labels, the network is enforced to discover integral object extents. To encode the visual words, we devise two learning strategies to learn the codebook and empirically verify their efficacy.
  \item We propose Hybrid Pooling (HP), a simple yet effective pooling approach, which incorporates local discriminative information and global information to aggregated less background and more object regions.
  \item By incorporating the proposed VWL and HP, we present a new classification network to generate CAMs with higher quality for the WSSS task. Our method achieves new state-of-the-art performance, \ie 70.6\% mIoU on PASCAL VOC 2012 $val$ set and 36.2\% mIoU on MS COCO 2014 $val$ set.
\end{itemize}

\par This paper is an improved version of our preliminary work \citep{ru2021learning}. Compared with the conference version, this work further improves the learning-based strategy and proposes the memory-bank strategy which could learn visual words better. The performance is remarkably improved with these improvements and surpasses the latest state-of-the-art methods. We also conduct further experiments on more datasets to verify the efficacy of our approach.

\par The rest of this paper is structured as follows. In Section~\ref{related_work}, we briefly introduce some related works on WSSS with image-level labels and improvements on CAMs. The detailed methods are presented in Section~\ref{methods}. We present the experimental settings and results in Section~\ref{experiments}. Section~\ref{conclusion} concludes our work.

\section{Related Work}
\label{related_work}

\subsection{WSSS with Image-level Labels}
\par Weakly-Supervised Semantic Segmentation (WSSS) aims to develop semantic segmentation models with weak annotations, such as image-level labels \citep{papandreou2015weakly,pinheiro2015image,ahn2018learning,lee2021anti}, points \citep{bearman2016s}, bounding boxes\citep{song2019box,oh2021background,lee2021bbam}, and scribbles \citep{lin2016scribblesup}. In this work, we focus on WSSS with only image-level labels. In the subsections below, we will introduce WSSS methods with image-level labels based on their motivations.

\paragraph{\textbf{Growing Seed Regions with Constraints.}} \citep{kolesnikov2016seed} proposed $SEC$ principle to expand the initial seed cues and coincide with the object shapes. This principle was adopted by subsequent works. For example, \citep{roy2017combining} used CRF-CNN \citep{zheng2015conditional} to refine the initial labels with low-level pixel information to generate pseudo labels fitting object boundaries. \citep{huang2018weakly} integrated seeded region growing \citep{adams1994seeded} to expand the initial seed cues generated from classification networks and also adopted dense CRF \citep{krahenbuhl2011efficient} to refine pseudo labels.

\paragraph{\textbf{Erasing.}} Based on the common observation that CAMs usually only captured the most discriminative regions, Wei \etal proposed to adversarially erase the discriminative regions and progressively localize the integral object regions \citep{wei2017object}. Similarly, ACoL \citep{zhang2018adversarial} used two parallel CNN to erase the feature maps in one branch with the discriminative regions derived from the other branch and fused the localized regions from both branches as outputs. To prevent the attention regions from shifting to non-object regions during erasing, SeeNet \citep{hou2018self} used saliency prior \citep{hou2017deeply} to suppress the attention in background regions.

\paragraph{\textbf{Accumulating Attention.}} Another interesting observation is that classification networks tend to shift attention to the different regions of the object across the training process \citep{jiang2019integral}. Motivated by this, Jiang \etal proposed OAA, which accumulated the activated regions during the different training stage. In \citep{yao2021non}, Yao \etal proposed a graph reasoning and non-salient region mining module to capture more object extents from non-salient regions, since the saliency prior used in OAA did not always correspond to the foreground objects. Kim \etal in \citep{kim2021discriminative} combined the idea of erasing and accumulating to suppress the discriminative regions in training, which could assist in finding less discriminative object regions.

\paragraph{\textbf{Mining Objects from Multiple Images.}} The works above mainly focused on mining semantic objects from single image. Some recent works also tried to leverage the semantic co-occurrence in two or more images \citep{fan2020cian,li2020group,sun2020mining}. In \citep{fan2020cian}, CIAN designed a cross image attention module to model the pixel-level affinity from different images with common semantics. In \citep{sun2020mining}, in addition to the co-attention from image pairs, Sun \etal further proposed a contrastive attention module that could mine the unique semantic objects. \citep{li2020group} used a graph neural network (GNN) \citep{scarselli2008graph} based approach to reason and capture the integral object information from a group of input images.

\paragraph{\textbf{Refining Seed Regions.}} Since CAMs only yield very coarse pixel labels, some pixel affinity based methods are proposed to refine CAMs and proved to work brilliantly. \citep{wang2020weakly} proposed an EM framework, in which a unary network was used to predict the class score maps and a pairwise network was used to learn the pixel affinities. The learned pixel affinity would be used to refine the score maps and then supervise the training of the framework. PSA \citep{ahn2018learning} derived foreground and background regions with high confidence from coarse labels and utilized the reliable regions to learn an affinity network. For each input image, their coarse labels were refined via random walk propagation \citep{vernaza2017learning} with the learned affinity matrix from the trained network. In further, IRNet \citep{ahn2019weakly} proposed to derive instance labels from instance-agnostic CAMs via additionally learning semantic instance boundaries and propagating the initial CAMs with the learned pixel affinities and instance boundaries.

\paragraph{\textbf{End-to-End Solutions.}} Though the majority of methods adopted a multi-step framework, some works also tried to devise elegant end-to-end models for WSSS with image-level labels. In \citep{papandreou2015weakly}, Papandreou \etal proposed an EM framework that estimated the pseudo labels with a probabilistic model and utilized the pseudo labels to train the network at the maximization step. \citep{zhang2020reliability} followed a similar framework of \citep{papandreou2015weakly} but leveraged CRF to refine CAMs and jointly minimized the cross-entropy loss and low-level energy loss with the highly-confident pseudo labels. \citep{araslanov2020single} devised normalized global weighted pooling to aggregate classification scores from predicted score maps, which could improve the completeness of objects. The predicted score maps were then refined with a pixel affinity-based module to supervise the training process. To avoid the network degrading to trivial solutions, \citep{araslanov2020single} additionally proposed a stochastic gate that randomly transferred low-level features to high-level semantics.

\subsection{Improvements on CAMs}
\par A prevailing series of WSSS with image-level labels is to derive better CAMs from the classification network. Many works have been proposed to produce better CAMs by encouraging more object extents to be discovered. In \citep{wang2020self}, Wang \etal observed that CAMs of the same image with different scaling ratio usually differed largely in shape, while they were supposed to be the same since they consisted of the same objects. They proposed to regularize the classification network by minimizing the difference between the CAMs of different scales and achieved remarkable performance. In \citep{chang2020weakly}, Chang \etal proposed to cluster the original semantic categories to sub-categories and further leveraged the sub-category labels to supervise the training of the network, which could enforce the network to discover more object regions to distinguish sub-categories. \citep{lee2021anti} proposed an anti-adversarial attack approach to gradually pull image away from decision boundaries which helped to discover more object regions. \citep{zhang2020causal} firstly introduced causal inference \citep{rubin2019essential} to alleviate the confounding bias attributed by ambiguous boundaries. Some recent works showed that data augmentation techniques \citep{chang2020mixup} and auxiliary self-supervised tasks \citep{jo2021puzzle} could help to discover more object regions and thus improve the quality of CAMs. In this work, we also focus on deriving better CAMs from classification networks but from two aspects. Specifically, we propose the visual words learning module and hybrid pooling approach to counter the problem of partial discriminative object regions unexpected background regions in CAMs.


\section{Methods}
\label{methods}

\par This section expatiates our proposed methods, including the Visual Words Learning (VWL) module, Hybrid Pooling (HP) approach, and training process of our network. 

\subsection{Method Overview}
\par
\begin{figure}[!tbp]
  \centering
  \includegraphics[width=0.48\textwidth]{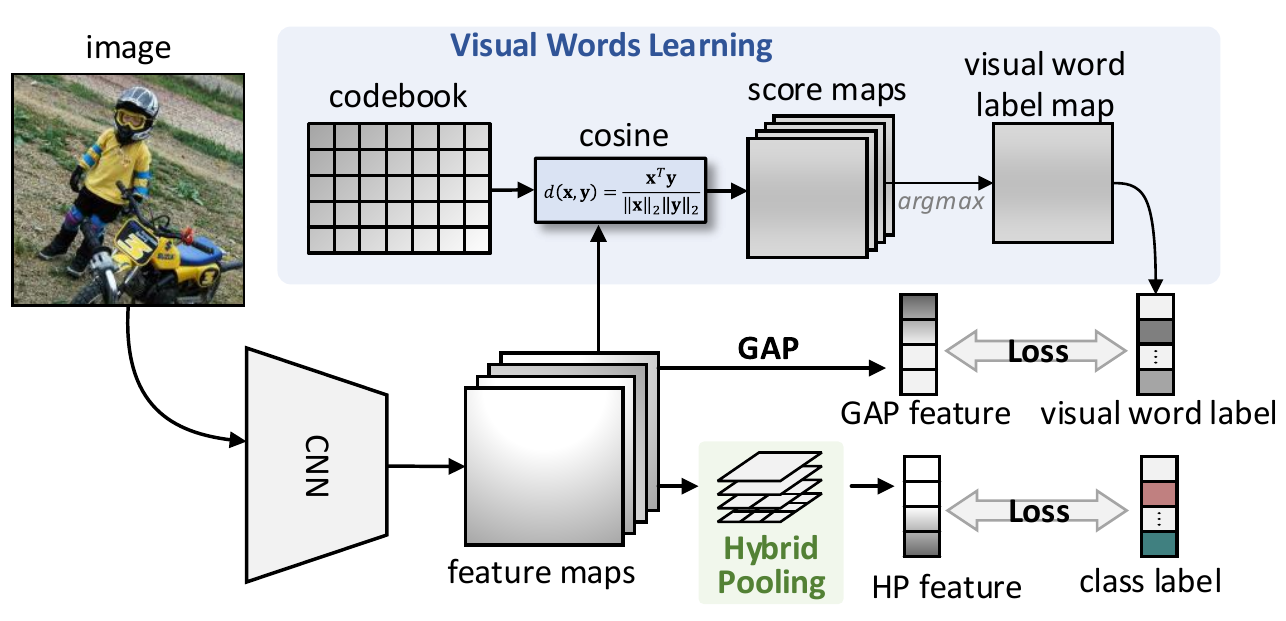}
  \caption{Overview of the proposed method. To encourage more object extents to be discovered, we propose the Visual Words Learning (VWL) module, which utilizes a codebook to encode the feature maps extracted by CNN. The encoded visual word labels are then used to supervise the training process of the classification network. We also propose a novel feature aggregation method, \ie Hybrid Pooling (HP), which incorporates GMP to reduce background information and GAP to ensure the object completeness in the generated CAMs.}
  \label{fig_overview}
\end{figure}

 The overall architecture of our method is presented in Fig.~\ref{fig_overview}. For an input image, we firstly use a CNN backbone to extract the convolutional feature maps. In the visual words learning module, a predefined codebook is employed to encode the feature maps to visual word score maps, in which each element denotes the probability of a pixel belonging to each visual word. The visual word label of a given image is derived from the score maps and further used to supervise the network training to activate more object regions. To alleviate the problem of unexpected background regions, we use the proposed hybrid pooling, which aggregates local discriminative information and global average information, so that less background is preserved in the generated CAMs.

\subsection{Preliminaries}
\par Currently, the majority of WSSS methods with image-level labels infer CAMs \citep{zhou2016learning} from a trained classification network as the initial coarse labels. In this work, we follow the original way in \citep{zhou2016learning} to directly produce CAMs with the feature maps of the last $conv$ layer and the weight matrix $\mathbf{W}^{img}$ in prediction layer. Specifically, CAMs for class $c$ are given by weighting each feature map in $\mathbf{F}$ with its contribution to class $c$
\begin{equation}
  \label{eq_cam_img}
  \mathbf{M}_c = \sum_{i=1}^d (\mathbf{W}^{img}_{i,c}\mathbf{F}_{:,:,i}).
\end{equation}

$\mathbf{M}_c$ is further passed through a $relu$ layer to eliminate the negative values, denoted as $\hat{\mathbf{M}}_c$. The generated $\hat{\mathbf{M}}_c$ will be used to produce pseudo segmentation labels with a background score threshold.

\subsection{Visual Words Learning}
\par As aforementioned, classification networks guided by image-level labels usually only discover partial discriminative extents of objects. The reason is that focusing on partial discriminative extents of objects is more beneficial to recognize different semantic categories. To solve this problem, our motivation is that if the network could be supervised with more fine-grained labels in the training procedure, it will be enforced to activate more semantic regions so that the generated CAMs would be more accurate.
To this end, we propose to jointly learn the visual words and image-level labels in the training process of classification networks.
\par Since only image-level labels are available in our task, in order to leverage visual word labels to guide the training of classification networks, we design an unsupervised visual words learning module. As shown in Fig.~\ref{fig_overview}, in the visual words learning module, a matrix $\mathbf{C}\in \mathbb{R}^{k\times d}$ is defined as the codebook, where $k$ is the number of words and $d$ denotes the feature dimension. $\mathbf{C}$ is utilized to encode the extracted convolutional feature map $\mathbf{F}\in \mathbb{R}^{h\times w \times d}$ to specific visual words. Here, we use $\cos$ distance to measure the similarity between the pixel at position $i$ in $\mathbf{F}$ and the $j$-th word in $\mathbf{C}$. The similarity matrix $\mathbf{S}$ is given as:
\begin{equation}
  \label{eq_s_ij}
  \mathbf{S}_{ij} = \frac{\mathbf{F}_i^\top \mathbf{C}_j}{||\mathbf{F}_i||_2||\mathbf{C}_j||_2}, 1 \leq i \leq hw, 1\leq j\leq k.
\end{equation}
\par The obtained $\mathbf{S}$ will be normalized row-wisely using $softmax$ to compute the probability of the $i$-th pixel in $\mathbf{F}$ belonging to $j$-th word in codebook $\mathbf{C}$. The process is given as
\begin{equation}
  \label{eq_p_ij}
  \mathbf{P}_{ij} =\frac{\exp(\tau \cdot \mathbf{S}_{ij})}{\sum_{n=1}^k \exp (\tau \cdot \mathbf{S}_{in})},
\end{equation}
where $\tau \textgreater 0$ is a temperature parameter to control the smoothness of $\mathbf{P}$.
\par The visual word label $\mathbf{Y}_{i}$ for $\mathbf{F}_i$ is then given as the word with the maximum probability, \ie, the index of the maximum value in the $i$-th row of $\mathbf{P}_{ij}$, which is denoted as
\begin{equation}
  \label{eq_y_ij}
  \mathbf{Y}_{i} = \mathop{\arg\max_{j} \mathbf{P}_{ij}}.
\end{equation}
\par For an input image $\mathbf{X}$, its visual word labels are computed as a $k$-dimensional vector $\mathbf{y}^{word}$, where $\mathbf{y}_j^{word}=1$ if the $j$-th word exists in $\mathbf{Y}$, and $\mathbf{y}_j^{word}=0$, otherwise. $\mathbf{y}^{word}$ will be used to guide the training procedure of the classification network to enforce it to discover more discriminative extents.
\par
\begin{figure*}
  \centering
  \begin{subfigure}[b]{0.48\textwidth}
    \centering
    \includegraphics[width=0.8\textwidth]{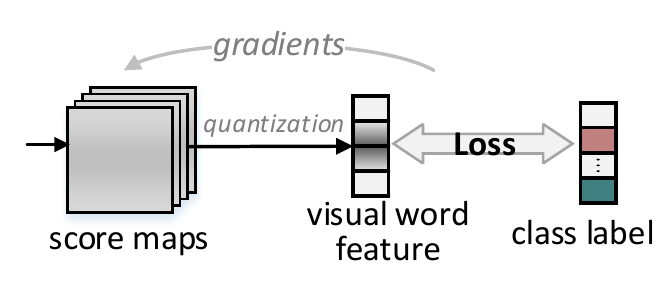}
    \caption{Learning-based strategy}
  \end{subfigure}
  \begin{subfigure}[b]{0.48\textwidth}
    \centering
    \includegraphics[width=0.95\textwidth]{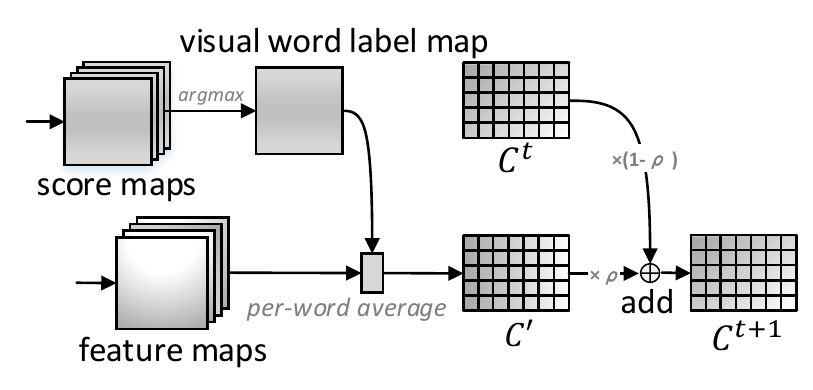}
    \caption{Memory-bank strategy}
  \end{subfigure}

  \caption{Illustration of strategies for visual word codebook. For the learning-based strategy, we follow the original intention of BoVW models \citep{liu2019bow,gidaris2020learning}, \ie using visual word frequencies to predict the image-level labels, which could enforce to learn the codebook from the back-propagated gradients. For the memory-bank strategy, inspired by mini-batch K-means \citep{sculley2010web}, we decompose the clustering on the whole dataset to each training step and update the codebook from reconstruction in the memory-bank mechanism.}
  \label{fig_strategy}
\end{figure*}

 Another problem is to ascertain the codebook for encoding visual word labels reasonably. In a classic BoVW model, the codebook is usually identified as the clustering centroids of the feature representations extracted from all local visual words \citep{liu2019bow,gidaris2020learning}. However, in our model, the feature representations for visual words are updated online as the training procedure. Therefore, the codebook $\mathbf{C}$ should also be updated online. To this end, as shown in Fig.~\ref{fig_strategy}, we devised two strategies to learn $\mathbf{C}$, namely the \textit{Learning-based strategy} and \textit{Memory-bank strategy}.

\paragraph{\textbf{Learning-based strategy.}}
\label{strategy_l}
\par In the learning-based strategy, following \citep{passalis2017learning,arandjelovic2017netvlad}, we set the codebook as a trainable parameter to learn it from back-propagated gradients. In a BoVW model, the frequencies of visual words are collected as the feature descriptor to predict the image classes so as to learn the relations between visual words and semantic classes \citep{liu2019bow}. However, this hard quantization approach will introduce non-continuities and is proved to make the training process intractable \citep{passalis2017learning}. In this work, we compute the frequency of each word by accumulating the probabilities in $\mathbf{P}$. Therefore, the soft frequency assignment of the $j$-th word is
\begin{equation}
  \label{eq_f_word}
  \mathbf{f}^{word}_j = \frac{1}{hw}\sum_{i=1}^{hw} \mathbf{P}_{ij},
\end{equation}
where $\mathbf{f}^{word}_j$ denotes the frequency of the $j$-th word in $\mathbf{F}$. As shown in Fig.~\ref{fig_strategy} (a), $\mathbf{f}^{word}$ will be used to predict the image-level labels, \ie, modeling the relations between visual words and image-level labels, which encourages the codebook to learn latent visual word representations via gradients.

\paragraph{\textbf{Memory-bank strategy.}}
\par As aforementioned, a classic BoVW model usually takes the clustering centroids of image features as the codebook. However, clustering on the whole dataset with the network training is extremely time-consuming. Inspired by mini-batch K-Means \citep{sculley2010web}, which decomposes clustering on a large-scale dataset to mini-batch iterations, we propose the {memory-bank strategy} to gradually update the codebook with reconstructed codebook in each training step.
\par In Eq.~\eqref{eq_y_ij}, the visual word label $\mathbf{Y}$ for each pixel in $\mathbf{F}$ is computed. We firstly transform $\mathbf{Y}\in \mathbb{R}^{hw}$ to $\mathbf{W}\in \mathbb{R}^{hw\times k}$ via one-hot encoding. The reconstructed codebook $\mathbf{C}'$ is hereby given as averaging the representations that are encoded to the same visual word category.
\begin{equation}
  \label{eq_c_re}
  \mathbf{C}'=\mathbf{D}^{-1}\mathbf{W}^\top \mathbf{F},
\end{equation}
where $\mathbf{D}$ is a degree matrix with $\mathbf{D}_{ii}={1}/{\sum_{j,k=i}\mathbf{W}_{jk}}$ and $\mathbf{D}_{ij}=0$ for off-diagonal elements. Here, we also assume $\mathbf{F}$ is unfolded to $\mathbb{R}^{hw\times d}$ for simplicity. As shown in Fig.~\ref{fig_strategy} (b), the reconstructed codebook $\mathbf{C}'$ is then used to update codebook with a momentum parameter $\rho$. This process is given as
\begin{equation}
  \label{eq_c_update}
  \mathbf{C}^{t+1}\leftarrow\rho\mathbf{C}'+(1-\rho)\mathbf{C}^t,
\end{equation}
where $t$ denotes the training iterations.

\subsection{Hybrid Pooling}
\begin{figure}[htbp]
  \centering
  \includegraphics[width=0.48\textwidth]{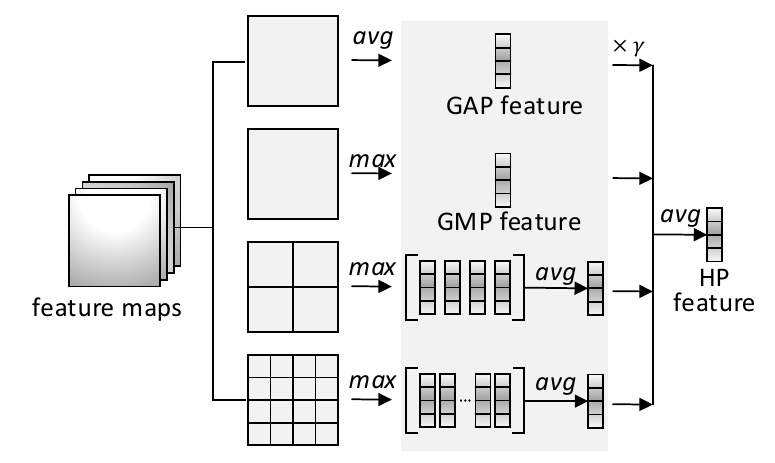}
  \caption{Illustration of the proposed hybrid pooling. $avg$: average pooling, $max$: max pooling.}
  \label{fig_hp}
\end{figure}
\par To mitigate the aforementioned disadvantages of GAP and GMP, in this work, we present a simple yet empirically effective pooling method which aggregates local maximum and global average values of the feature maps.

\par Considering the output feature map $\mathbf{F}$ with size of $h \times w \times d$ of the last convolutional layer, we firstly partition $\mathbf{F}$ to multi-scale divisions. As illustrated in Fig.~\ref{fig_hp}, each division with size of $\frac{h}{r}\times\frac{w}{r}\times d$ is pooled to a $d$-dimensional vector via $\mathbf{max}$ pooling, where $r\in\{1,2,4\}$ denotes the split size. $\mathbf{F}$ is thus aggregated to $\mathbf{F}^{max}$ with size of $r\times r\times d$. It is conspicuous that $\mathbf{F}^{max}$ only involves local maximum pixels so that less background is considered. We then pool $\mathbf{F}^{max}$ for subsequent classification task. The pooled feature $\mathbf{f}^{max}_r$ with split size $r$ is given by
\begin{equation}
  \label{eq_fmax}
  \mathbf{f}^{max}_r = \frac{1}{r^2}\sum_{i=1}^r\sum_{j=1}^r\mathbf{F}^{max}_{i,j,:}.
\end{equation}

\par Note that $\mathbf{f}^{max}_r$ only preserves the maximum responses of local objects, which may corrupt the completeness of objects. To tackle this problem, in HP module, we also incorporate the results of GAP. Given the pooled feature of GAP layer, which is computed as
\begin{equation}
  \label{eq_fgap}
  \mathbf{f}^{gap} = \frac{1}{hw}\sum_{i=1}^h\sum_{j=1}^w\mathbf{F}_{i,j,:},
\end{equation}
the final output of hybrid pooling module is calculated by weighting the outputs in Eq.~\eqref{eq_fmax} and Eq.~\eqref{eq_fgap}, computed as
\begin{equation}
  \label{eq_hp}
  \mathbf{f}^{hp} = \frac{1}{\gamma+3} (\sum_{r\in\{1,2,4\}}\mathbf{f}^{max}_r+ \gamma \mathbf{f}^{gap}),
\end{equation}
where $\gamma$ is a weight factor. Leveraging Eq.~\eqref{eq_hp}, more regions of foreground objects and less background are captured for classification, so that the generated CAMs could coincide better with object shapes.

\subsection{Network Training}
\label{sec_net_training}
\par Since only image-level labels are available, the classification loss is indispensable to train the network. After calculating $\mathbf{f}^{hp}$ via Eq.~\eqref{eq_hp}, the classification score for image label is computed with a classification layer (an $1\times 1\ conv$ layer in practice), denoted as $\mathbf{p}^{img}=conv(\mathbf{f}^{hp},\mathbf{W}^{img})$, where $\mathbf{W}^{img}$ is the weight matrix of this layer. As a common practice \citep{wang2020self,chang2020weakly,araslanov2020single}, the multi-label soft margin loss \citep{paszke2019pytorch} is employed to compute the classification loss
\begin{equation}
  \label{eq_loss_img}
  \begin{split}
    \mathcal{L}_{cls}(\mathbf{p}^{img},\mathbf{y}^{img}) &= \frac{1}{L}\sum_{i=1}^L[\mathbf{y}_i^{img}\log\frac{\exp(\mathbf{p}^{img}_i)}{1+\exp(\mathbf{p}^{img}_i)}\\
    &+(1-\mathbf{y}_i^{img})\log\frac{1}{1+\exp(\mathbf{p}^{img}_i)}],
  \end{split}
\end{equation}
where $\mathbf{y}^{img}$ denotes the ground-truth image label and $L$ is the number of image classes.

\par To capture more semantic regions, the pooled feature is also used to predict the visual word label $\mathbf{y}^{word}$ generated in previous steps. It is noted that $\mathbf{y}^{word}$ is generated based on all pixels in feature map $\mathbf{F}$. Therefore, we use GAP here instead of our HP to perform feature aggregation for predicting $y^{word}$. The predicted visual word score is thus denoted as $\mathbf{p}^{word}=conv(\mathbf{f}^{gap},\mathbf{W}^{word})$, where $\mathbf{W}^{word}$ is the weight matrix of the prediction layer. The classification loss for visual words is then denoted as $\mathcal{L}_{cls}(\mathbf{p}^{word},\mathbf{y}^{word})$, which is in the same form as Eq.~\eqref{eq_loss_img}. The overall loss function is the sum of $\mathcal{L}_{cls}(\mathbf{p}^{img},\mathbf{y}^{img})$ and $\mathcal{L}_{cls}(\mathbf{p}^{word},\mathbf{y}^{word})$, namely
\begin{equation}
  \label{eq_loss_all_1}
  \begin{split}
    \mathcal{L}_{all} = \mathcal{L}_{cls}(\mathbf{p}^{img},\mathbf{y}^{img})+\mathcal{L}_{cls}(\mathbf{p}^{word},\mathbf{y}^{word}).
  \end{split}
\end{equation}
\paragraph{\textbf{Auxiliary Loss for Learning-based Strategy.}} Recall that in the learning-based strategy, to learn the codebook from gradients, we set the codebook as a trainable parameter and leverage the visual word representations to learn the image-level labels. Specifically, the visual word frequency $\mathbf{f}^{word}$ acquired in Eq.~\eqref{eq_f_word} is projected into the class probability space with an $1\times 1\ conv$ layer. The predicted score is denoted by $\mathbf{p}^{w2i}$ so that the loss function is given as $\mathcal{L}_{cls}(\mathbf{p}^{w2i},\mathbf{y}^{img})$.
\par We empirically found that learning with the loss $\mathcal{L}_{cls}(\mathbf{p}^{w2i},\mathbf{y}^{img})$ solely tended to make the learned visual word representations in codebook $\mathbf{C}$ redundant. To tackle this problem, we add a regularization term to minimize the correlation between different latent visual word representations. Here, we use the DeCov loss \citep{cogswell2015reducing}, which reduces the correlations between rows in a matrix by minimizing their covariance, \ie
\begin{equation}
  \label{eq_loss_decov}
  \begin{split}
    \mathcal{L}_{decov} = \frac{1}{2}(\|\mathbf{\hat{C}}\|_F^2-\|\text{diag}(\mathbf{\hat{C}})\|_2^2),
  \end{split}
\end{equation}
where $\mathbf{\hat{C}}$ is the covariance matrix of $\mathbf{C}$, $\|\cdot\|_F$ denotes the Frobenius norm, and $\text{diag}(\cdot)$ extracts the main diagonal elements of a matrix to a vector.
\par The auxiliary loss for the learning-based strategy is then given as the sum of aforementioned two losses
\begin{equation}
  \label{eq_loss_aux}
  \begin{split}
    \mathcal{L}_{aux} = \mathcal{L}_{cls}(\mathbf{p}^{w2i},\mathbf{y}^{img})+\mathcal{L}_{decov} .
  \end{split}
\end{equation}
\par By regularizing Eq.~\eqref{eq_loss_all_1} with Eq.~\eqref{eq_loss_aux}, we present the overall loss function to optimize the network under the learning-based strategy
\begin{equation}
  \label{eq_loss_all_2}
  \begin{split}
    \mathcal{L}_{all} = \mathcal{L}_{cls}(\mathbf{p}^{img},\mathbf{y}^{img})+\mathcal{L}_{cls}(\mathbf{p}^{word},\mathbf{y}^{word})+\mathcal{L}_{aux}.
  \end{split}
\end{equation}
\par The overall training process with learning or memory-bank strategy is summarized in Alg.~\ref{alg_learn}.

\SetKwInput{KwParam}{Params}
\begin{algorithm}
  \SetAlgoLined
  \KwIn{Image $\mathbf{I}$, label $\mathbf{y}^{img}$;}
  \KwParam{Backbone network $\mathbb{E}(\cdot,\theta)$, codebook $\mathbf{C}$, hyper-parameters $\{k, \gamma, \tau, \rho\}$;}
  Initialize $\mathbb{E}(\cdot,\theta)$ and $\mathbf{C}$\;
  \While{training}{
    Extract feature maps $\mathbf{F}=\mathbb{E}(\mathbf{I},\theta)$\;
    Compute visual word label $\mathbf{y}^{word}$ via Eq.~\eqref{eq_s_ij} to Eq.~\eqref{eq_y_ij}\;
    Compute pooling features via Eq.~\eqref{eq_fgap} and Eq.~\eqref{eq_hp}\;

    \eIf{Learning-based strategy}{
      Compute visual word feature via Eq.~\eqref{eq_f_word}\;
      Compute loss via Eq.~\eqref{eq_loss_all_2}\;
      Optimize $\mathbb{E}(\cdot,\theta)$ and $\mathbf{C}$;
    }(\textit{Memory-bank strategy}){
      Compute loss via Eq.~\eqref{eq_loss_all_1}\;
      Optimize $\mathbb{E}(\cdot,\theta)$\;
      Reconstructing $\mathbf{C}'$ via Eq.~\eqref{eq_c_re}\;
      Update $\mathbf{C}\leftarrow\rho\mathbf{C}'+(1-\rho)\mathbf{C}$\;
    }
  }
  \caption{Training procedure of the proposed network.}
  \label{alg_learn}
\end{algorithm}

\section{Experiments}
\label{experiments}

\subsection{Implementation Details}
\paragraph{\textbf{Dataset.}}
\par We evaluated our method on the PASCAL VOC 2012 dataset \citep{everingham2010pascal} and MS COCO 2014 dataset \citep{lin2014microsoft}. For all experiments, the mean Intersection-over-Union (mIoU) ratio was used as the evaluation criteria.
\par {PASCAL VOC 2012 dataset} \citep{everingham2010pascal} consists of 21 semantic categories, including 20 foreground object classes and the background class. Following the common practice\citep{chen2017deeplab,wang2020self,chang2020weakly}, this dataset is augmented with SBD dataset \citep{hariharan2011semantic}. The $train$, $val$, and $test$ set of the augmented dataset consist of 10,582, 1449, and 1456 images, respectively.
\par {MS COCO 2014 dataset} \citep{lin2014microsoft} is a large-scale dataset with 81 semantic categories, including the background class. After excluding the images without annotations \citep{lee2021railroad}, the MS COCO dataset consists of 82,081 and 40,137 images in $train$ and $val$ set, respectively.

\begin{table}[!tp]

  \centering
  \caption{Evaluation and comparison of the generated CAMs and pseudo labels in mIoU. The best results are highlighted in \textbf{bold}.}
  \label{tab_voc_cam}
  \scalebox{1}{
    \begin{tabular}{l|ccc}
      \toprule
      {Method}                     & {CAMs}        & $+CRF$        & $+Ref.$       \\\midrule
      \multicolumn{4}{l}{\textit{CAMs refined with PSA} \citep{ahn2018learning}.}  \\
      PSA    \tiny CVPR'2018       & 48.0          & --            & 61.0          \\
      Mixup CAM \tiny BMVC'2020    & 50.1          & --            & 61.9          \\
      SC-CAM  \tiny CVPR'2020      & 50.9          & 55.3          & 63.4          \\
      SEAM       \tiny CVPR'2020   & 55.4          & 56.8          & 63.6          \\
      PuzzleCAM   \tiny arXiv'2021 & 51.5          & --            & 64.7          \\
      AdvCAM      \tiny CVPR'2021  & 55.6          & 62.1          & 68.0          \\\midrule
      \multicolumn{4}{l}{\textit{CAMs refined with IRNet} \citep{ahn2019weakly}.}  \\
      IRNet   \tiny CVPR'2019      & 48.8          & 54.3          & 66.3          \\
      MBMNet   \tiny ACM MM'2020   & 50.2          & --            & 66.8          \\
      CDA      \tiny arXiv'2021    & 50.8          & --            & 67.7          \\
      VWE   \tiny IJCAI'2021       & 55.1          & 60.9          & 69.5          \\
      CONTA \tiny NeurIPS'2020     & 56.2          & --            & 67.9          \\
      AdvCAM  \tiny CVPR'2021      & 55.6          & 62.1          & 69.9          \\
      \rowcolor[HTML]{fafafa}
      {Ours-M}                     & 56.9          & 62.6          & 71.1          \\
      \rowcolor[HTML]{eaeaea}
      {Ours-L}                     & \textbf{57.3} & \textbf{63.0} & \textbf{71.4} \\ \bottomrule
    \end{tabular}}

\end{table}

\paragraph{\textbf{Classification Network.}}
\par For the network to produce CAMs, we used ResNet101 \citep{he2016deep} pre-trained on ImageNet \citep{krizhevsky2012imagenet} as the backbone to extract convolutional feature maps. For PASCAL VOC and MS COCO datasets, the classification network was trained for 6 epochs, with a batch size of 16. To optimize the network, we used the SGD optimizer with momentum mechanism and set the momentum coefficient as 0.9. The learning rate was initially set to 0.01 for the backbone parameters and 0.1 for the other parameters. All learning rates were decayed every iteration with a polynomial decay scheduler. Specifically, in each iteration, the learning rate was multiplied by $(1-\frac{iter}{max\_iter})^{power}$, with $power=0.9$.

\par As for the hyper-parameters, the number of visual words $k$ and the weight factor $\gamma$ in Eq.~\eqref{eq_hp} were set to 256 and 2, respectively. The temperature parameter $\tau$ in Eq.~\eqref{eq_p_ij} was empirically set to 1. For the memory-bank strategy, we had an extra momentum coefficient $\rho$ in Eq.~\eqref{eq_c_update}, which was set to 0.001. More details and the impacts of these hyper-parameters are reported in Section~\ref{hyper_params}. Our code is available at \url{https://github.com/rulixiang/vwe/tree/master/v2}.

\paragraph{\textbf{CAMs Refinement.}}
\par The generated initial labels by directly segmenting CAMs with thresholds are usually very coarse \citep{ahn2018learning,ahn2019weakly}. To improve the quality of pseudo labels and the semantic segmentation performance, we adopted IRNet \citep{ahn2019weakly} as the refinement approach for processing the initial coarse labels generated from the classification network. In practice, we used the official implementation\footnote{\url{https://github.com/jiwoon-ahn/irn}} without changing their settings.

\paragraph{\textbf{Segmentation Network.}}
\par For the semantic segmentation network, we used the DeepLabV2 \citep{chen2017deeplab} system with ResNet101 \citep{he2016deep} as backbone, which is a prevailing choice for WSSS task \citep{ke2021universal,lee2021anti,chang2020weakly}. For experiments on PASCAL VOC 2012 dataset \citep{everingham2010pascal}, we followed the default settings of DeepLabV2 \citep{chen2017deeplab}, \ie, the learning rate was initially set to 0.001 and decayed with a polynomial scheduler. The batch size and number of iterations were 10 and 20,000, respectively. We used a momentum optimizer with the momentum parameter of 0.9 and weights decay rate of 0.0005. For a fair comparison with other WSSS works, we evaluated the DeepLabV2 initialized with ImageNet \citep{krizhevsky2012imagenet} and MS COCO dataset \citep{lin2014microsoft} pre-trained weights. For experiments on the MS COCO dataset \citep{lin2014microsoft}, we followed the same settings as the experiments on the PASCAL VOC dataset. The only difference was that we trained the segmentation network for 60,000 iterations since MS COCO consisted of much more samples.

\subsection{Results on PASCAL VOC dataset}

\begin{figure*}[!tp]
  \centering
  \includegraphics[width=0.96\textwidth]{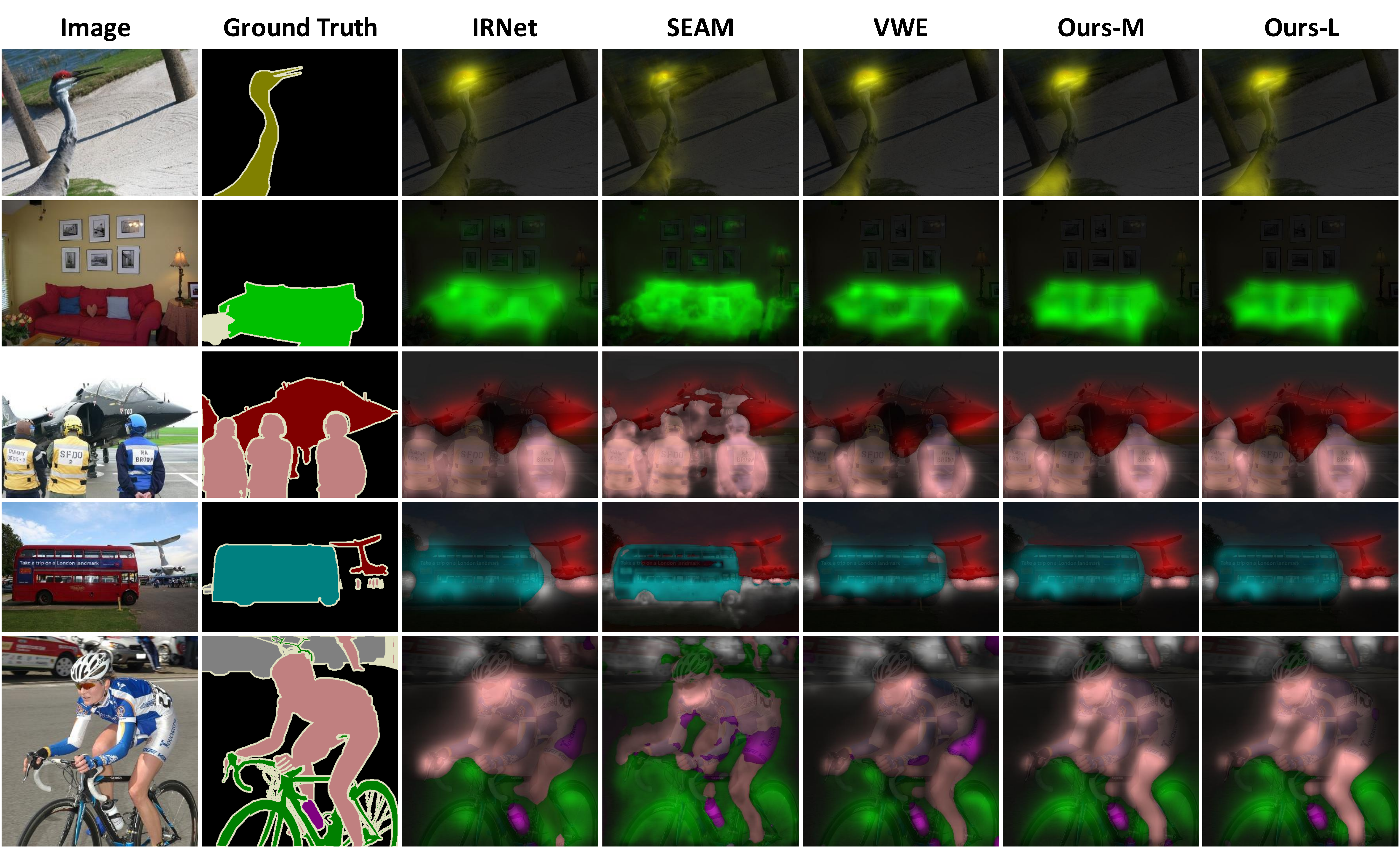}
  \caption{Visualization of the generated CAMs. Different colors denote the activated regions of different semantic categories.}
  \label{fig_vis_cam}
\end{figure*}

\begin{table}[!tp]

  \centering
  \caption{Semantic segmentation results on PASCAL VOC 2012 dataset. The best results are highlighted in \textbf{bold}. $Sup.$ denotes supervision type. $Seg.$ denotes segmentation network. }
  \label{tab_voc_seg}
  \setlength{\tabcolsep}{2pt}
  \begin{tabular}{l|c|c|cc}
    \toprule
    {Method}                                           & {$Sup.$}                          & \multicolumn{1}{c|}{ $Seg.$ } & {$val$}       & {$test$}                                                                                  \\  \midrule
    \multicolumn{5}{l}{\textit{Full Supervision}.}                                                                                                                                                                                     \\
    (1)$^\dagger$  DeepLabV1$^\dagger$ \tiny ICLR'2015 & \multirow{5}{*}{$\mathcal{F}$}    & \multicolumn{1}{c|}{ -- }     & 75.5          & --                                                                                        \\
    (2) DeepLabV2 \tiny TPAMI'2017                     &                                   & \multicolumn{1}{c|}{ -- }     & 76.3$^*$      & --                                                                                        \\
    (2)$^\dagger$ DeepLabV2$^\dagger$ \tiny TPAMI'2017 &                                   & \multicolumn{1}{c|}{ -- }     & 77.6          & 79.7                                                                                      \\
    (3) WideResNet38 \tiny PR'2019                     &                                   & \multicolumn{1}{c|}{ -- }     & 80.8          & 82.5                                                                                      \\
    (4) Res2Net101 \tiny TPAMI'2021                     &                                   & \multicolumn{1}{c|}{ -- }     & 80.2          & --                                                                                      \\\midrule
    \multicolumn{5}{l}{\textit{Image-level Supervision + Saliency Maps}.}                                                                                                                                                              \\
    OAA+ \tiny ICCV'2019                               & \multirow{8}{*}{$\mathcal{I+S}$} & (1)$^\dagger$                 & 65.2          & 66.4                                                                                      \\
    Li \etal \tiny AAAI'2021                           &                                   & (2)                           & 68.2          & 68.5                                                                                      \\
    NSROM \tiny CVPR'2021                              &                                   & (2)                           & 68.3          & 68.5                                                                                      \\
    NSROM \tiny CVPR'2021                              &                                   & (2)$^\dagger$                 & 70.4          & 70.2                                                                                      \\
    DRS \tiny AAAI'2021                                &                                   & (2)$^\dagger$                 & 70.4          & 70.7                                                                                      \\
    EPS \tiny CVPR'2021                                &                                   & (2)$^\dagger$                 & \textbf{70.9} & \textbf{70.8}                                                                             \\
    AuxSegNet \tiny ICCV'2021                          &                                   & (3)                           & 69.0          & 68.6                                                                                      \\
    EDAM \tiny CVPR'2021                               &                                   & (2)$^\dagger$                 & \textbf{70.9} & 70.6                                                                                      \\ \midrule
    \multicolumn{5}{l}{\textit{Image-level Supervision Only}.}                                                                                                                                                                         \\
    IAL \tiny IJCV'2020                                & \multirow{14}{*}{$\mathcal{I}$}   & (2)                           & 64.3          & 65.4                                                                                      \\
    SEAM \tiny CVPR'2020                               &                                   & (3)                           & 64.5          & 65.7                                                                                      \\
    A$^2$GNN \tiny TPAMI'2021                          &                                   & (2)                           & 66.8          & 67.4                                                                                      \\
    VWE \tiny IJCAI'2021                               &                                   & (2)$^\dagger$                 & 69.6          & 69.3                                                                                      \\
    AdvCAM \tiny CVPR'2021                             &                                   & (2)                           & 68.1          & 68.0                                                                                      \\
    OC-CSE \tiny ICCV'2021                             &                                   & (3)                           & 68.4          & 68.2                                                                                      \\
    ESCNet \tiny ICCV'2021                             &                                   & (3)                           & 66.6          & 67.6                                                                                      \\
    CDA \tiny ICCV'2021                                &                                   & (3)                           & 66.1          & 66.8                                                                                      \\
    CPN \tiny ICCV'2021                                &                                   & (3)                           & 67.8          & 68.5                                                                                      \\
    PMM \tiny ICCV'2021                                &                                   & (4)                           & 70.0          & 70.5                                                                                      \\
    \rowcolor[HTML]{fafafa}
                                                       &                                   & (2)                           & 68.7          & 69.2\tablefootnote{\url{http://host.robots.ox.ac.uk:8080/anonymous/XJDOJG.html}}          \\
    \rowcolor[HTML]{fafafa}
    \multirow{-2}{*}{Ours-M}                           &                                   & (2)$^\dagger$                 & \textbf{70.6} & 70.4\tablefootnote{\url{http://host.robots.ox.ac.uk:8080/anonymous/J00QBG.html}}          \\
    \rowcolor[HTML]{eaeaea}
                                                       &                                   & (2)                           & 69.2          & 69.2\tablefootnote{\url{http://host.robots.ox.ac.uk:8080/anonymous/Y0XECB.html}}          \\
    \rowcolor[HTML]{eaeaea}
    \multirow{-2}{*}{Ours-L}                           &                                   & (2)$^\dagger$                 & \textbf{70.6} & \textbf{70.7}\tablefootnote{\url{http://host.robots.ox.ac.uk:8080/anonymous/0QVYDO.html}} \\
    \bottomrule
    \multicolumn{4}{l}{\textit{$^*$ Accuracy obtained with our re-implementation}.}                                                                                                                                                    \\
    \multicolumn{4}{l}{\textit{$^\dagger$ Backbone pre-trained on MS COCO dataset}.}
  \end{tabular}

\end{table}
\par Table~\ref{tab_voc_cam} reports the quantitative evaluation results of CAMs on the $train$ set of the PASCAL VOC dataset. $CRF$ denotes the generated CAMs are refined with dense CRF \citep{krahenbuhl2011efficient}. $Ref.$ denotes the generated CAMs are refined with PSA \citep{ahn2018learning} or IRNet \citep{ahn2019weakly}. The best results are highlighted in bold. We denote our methods with memory-bank strategy and learning-based strategy as Ours-M and Our-L, respectively. Ours results are compared with recent related works on improving the quality of CAMs, including AdvCAM\citep{lee2021anti}, SC-CAM \citep{chang2020weakly}, CONTA \citep{zhang2020causal}, and SEAM \citep{wang2020self} \etc. Table~\ref{tab_voc_cam} shows that our methods with learning-based and memory-bank strategy could both remarkably surpass current state-of-the-art works. After further refinement with IRNet \citep{ahn2019weakly}, Our-M and Ours-L achieve 71.1\% and 71.4\% mIoU on the pseudo labels, respectively, which also outperform the competitors.

\par In Fig.~\ref{fig_vis_cam}, we visualize the generated CAMs and compare them with the results of recent methods, including IRNet \citep{ahn2019weakly}, SEAM \citep{wang2020self}, and VWE (our previous work with HP and simple visual words encoder) \citep{ru2021learning}. The results of the learning-based strategy (Ours-L) and the memory-bank strategy (Ours-M) are both presented. It is observed that our results typically activate more object regions and less mis-activated background, which is owing to that the proposed visual words learning module encourages to discover more objects, while HP aggregates local discriminative information and thereby reduces background regions. It is also noticed that the results of Ours-L and Ours-M are very close visually (Fig.~\ref{fig_vis_cam}) and numerically (Table~\ref{tab_voc_cam}), which indicates that both the learning-based and memory-bank strategy could work finely.

\par We use the refined CAMs as the pseudo labels to train regular semantic segmentation networks and compare the results on the $val$ and $test$ set of PASCAL VOC dataset. The results are reported in Table~\ref{tab_voc_seg}. For a fair comparison, we report the performance using DeepLabV2 with backbone pre-trained on ImageNet \citep{chen2017deeplab} and MS COCO \citep{lin2014microsoft}. By default, the presented results are obtained with dense CRF post-processing \citep{krahenbuhl2011efficient}. It is observed that, for the WSSS methods with only image-level labels, our method obtains the best performance. Specifically, Ours-L achieves 69.2\% and 70.6\% mIoU on the PASCAL VOC $val$ set with DeepLabV2 initialized with ImageNet and MS COCO pre-trained weights, respectively, which recover 90.7\% and 91.0\% of the upper bound of their fully-supervised counterparts. Our methods also achieve comparable performance with recent state-of-the-art WSSS methods using extra saliency maps, such as NSROM \citep{yao2021non}, DRS \citep{kim2021discriminative}, EPS \citep{lee2021railroad}, AuxSegNet \citep{xu2021leveraging}, and EDAM \citep{wu2021embedded}. Our method also outperforms recent methods with superior backbone networks, such as PMM \citep{li2021pseudo}, which uses Res2Net101 \citep{gao2021res2net} as the backbone for semantic segmentation. Note that both Ours-M and Ours-L could surpass recent WSSS methods with only image-level supervision, which demonstrates the efficacy of our proposed learning-based and memory-bank strategies.

\begin{figure*}[htp]
  \centering
  \includegraphics[width=0.98\textwidth]{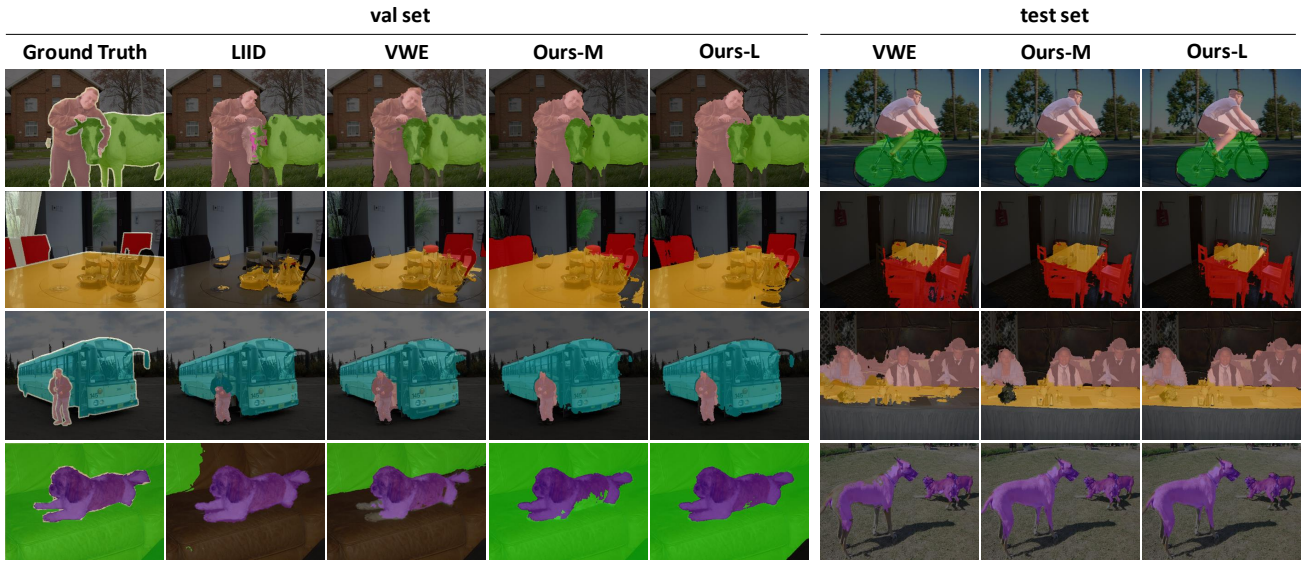}
  \caption{Examples of the predicted segmentation from PASCAL VOC $val$ and $test$ set.}
  \label{fig_vis_seg}
\end{figure*}
\par The qualitative results of our proposed method and some other methods' results, including LIID \citep{liu2020leveraging} and VWE \citep{ru2021learning}, are presented in Fig.~\ref{fig_vis_seg}. We could observe that our method significantly outperforms other WSSS methods and coincides better with the ground-truth labels.

\subsection{Results on MS COCO dataset}
\par To further verify the efficacy of the proposed method, we conduct experiments on the MS COCO dataset \citep{lin2014microsoft}, which consists of much more images and semantic categories than PASCAL VOC 2012 dataset. The quantitative evaluation results on the MS COCO dataset are presented in Table~\ref{tab_coco_seg}. We observe that Ours-L and Ours-M achieve 36.2\% and 36.1\% mIoU on the MS COCO $val$ dataset, respectively. Both of them could outperform other WSSS methods with only image-level labels. Besides, our results are also better than the results of recent state-of-the-art WSSS methods with image-level labels and extra saliency cues. The superiority of the performance on the MS COCO dataset also demonstrates the efficacy of our methods.

\begin{table}[!tp]

  \centering
  \caption{Semantic segmentation results on MS COCO dataset. The best results are highlighted in \textbf{bold}. $Sup.$ denotes supervision type. $Seg.$ denotes segmentation network. }
  \label{tab_coco_seg}
  \begin{tabular}{l|c|c|c}
    \toprule
    {Method}                               & $Sup.$                           & \multicolumn{1}{c|}{ $Seg.$ } & $val$         \\\midrule
    \multicolumn{4}{l}{\textit{Image-level Supervision + Saliency Maps}.}                                                     \\
    DSRG               \tiny CVPR'2018     & \multirow{5}{*}{$\mathcal{I+S}$} & DeepLabV2                     & 26.0          \\
    Li \etal               \tiny AAAI'2020 &                                  & DeepLabV2                     & 28.4          \\
    ADL            \tiny TPAMI'2020        &                                  & DeepLabV2                     & 30.8          \\
    EPS              \tiny CVPR'2021       &                                  & DeepLabV2                     & 35.7          \\
    AuxSegNet              \tiny ICCV'2021 &                                  & WideResNet38                  & 33.9          \\ \midrule
    \multicolumn{4}{l}{\textit{Image-level Supervision Only}.}                                                                \\
    SEC             \tiny ECCV'2016        & \multirow{9}{*}{$\mathcal{I}$}   & DeepLabV2                     & 22.4          \\
    Saleh \etal  \tiny TPAMI'2017          &                                  & DeepLabV2                     & 20.4          \\
    IAL               \tiny IJCV'2020      &                                  & DeepLabV2                     & 27.7          \\
    SEAM                  \tiny CVPR'2020  &                                  & WideResNet38                  & 31.9          \\
    CONTA          \tiny NeurIPS'2020      &                                  & WideResNet38                  & 32.8          \\
    CDA          \tiny ICCV'2021           &                                  & WideResNet38                  & 33.2          \\
    PMM              \tiny ICCV'2021       &                                  & Res2Net101                  & 35.7          \\
    \rowcolor[HTML]{fafafa}
    Ours-M                                 &                                  & DeepLabV2                     & 36.1          \\
    \rowcolor[HTML]{eaeaea}
    Ours-L                                 &                                  & DeepLabV2                     & \textbf{36.2} \\ \bottomrule
  \end{tabular}

\end{table}

\par
\begin{figure*}[htbp]
  \centering
  \includegraphics[width=0.99\textwidth]{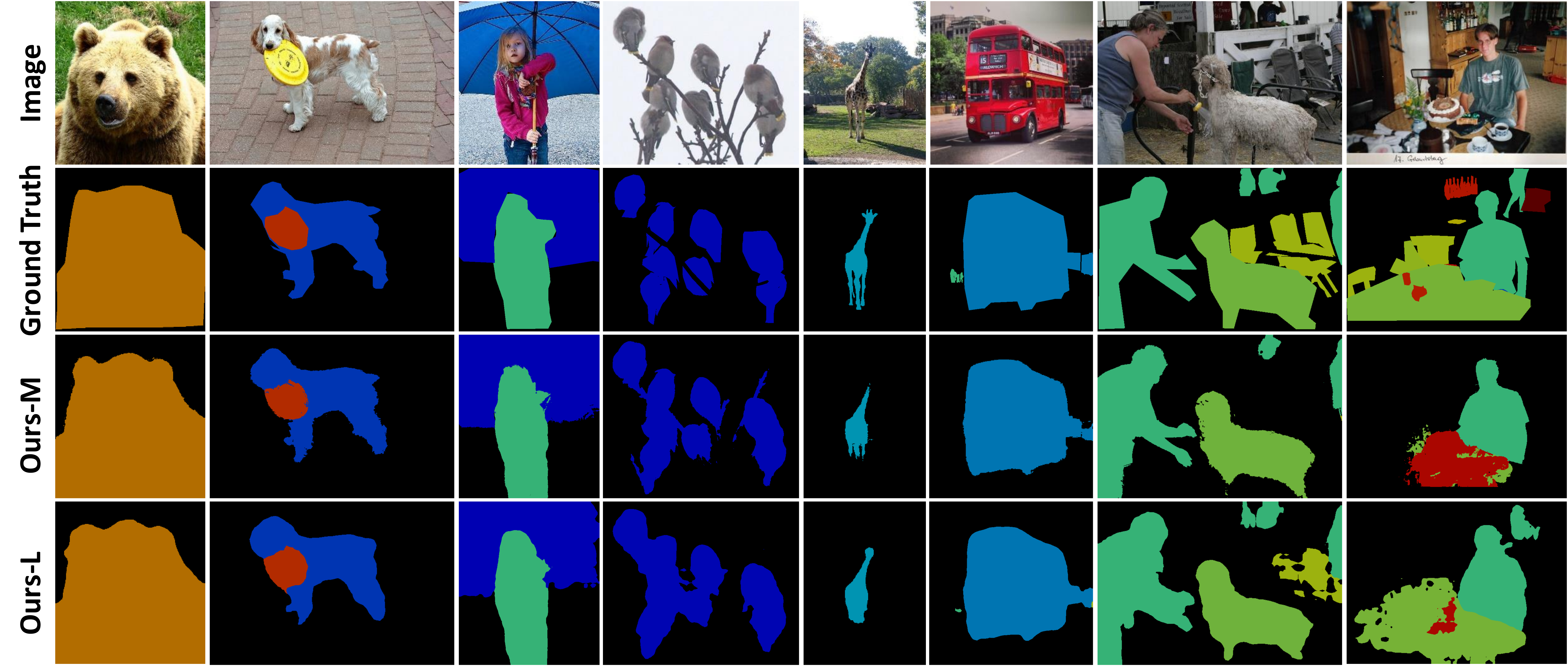}
  \caption{Examples of the predicted labels from MS COCO $val$ dataset.}
  \label{fig_vis_seg_coco}
\end{figure*}

 We present some predicted example images of MS COCO $val$ dataset in Fig.~\ref{fig_vis_seg_coco}. It is observed that our method could produce comparable results with ground-truth labels, though the MS COCO dataset is much more challenging. However, when the background is complex, as presented in the last two columns, the predicted results are clearly worse than the ground-truth labels.

\subsection{Ablation Study and Analysis}
\label{ablation}

\paragraph{\textbf{Quantitative Ablation Results.}} We conducted ablation experiments on the PASCAL VOC dataset to show the efficacy of the proposed methods. The quantitative evaluation results of the generated CAMs using baseline with different modules are presented in Table~\ref{tab_ablation}. VWE denotes the visual words learning module without DeCov regularization in our preliminary work \citep{ru2021learning}. VWL-M and VWL-L denote the proposed visual words learning module with learning-based and memory-bank strategy, respectively. We observe that the both proposed HP and VWL module could improve the quality of the generated CAMs. Besides, our proposed memory-bank and learning-based strategy could further improve the mIoU on the $train$ set to about 57\%, which remarkably outperform recent state-of-the-art methods presented in Table~\ref{tab_voc_cam}.

\begin{table}[t]

  \caption{Ablation studies of our proposed methods on the $train$ and $val$ set. The best results are highlighted in \textbf{bold}.}
  \label{tab_ablation}
  \small
  \centering
  \setlength{\tabcolsep}{2pt}
  \begin{tabular}{l|cccc|cc}
    \toprule
    Backbone                   & HP         & VWE        & VWL-M      & VWL-L      & $train$                     & $val$                       \\ \midrule
    ResNet50                   &            &            &            &            & 48.3                        & 47.0                        \\
    ResNet101                  &            &            &            &            & 49.5                        & 48.4                        \\ \midrule
    \multirow{4}{*}{ResNet101} & \checkmark &            &            &            & 54.0 {\tiny{+4.5}}          & 53.1 {\tiny{+4.7}}          \\
                               & \checkmark & \checkmark &            &            & 55.1 {\tiny{+5.6}}          & 54.8 {\tiny{+6.4}}          \\
                               & \checkmark &            & \checkmark &            & 56.9 {\tiny{+7.4}}          & 56.4 {\tiny{+8.0}}          \\
                               & \checkmark &            &            & \checkmark & \textbf{57.3} {\tiny{+7.8}} & \textbf{56.9} {\tiny{+8.5}} \\ \bottomrule
  \end{tabular}
\end{table}

\begin{figure}[htp]
  \centering
  \setlength{\tabcolsep}{2pt}
  \includegraphics[width=0.48\textwidth]{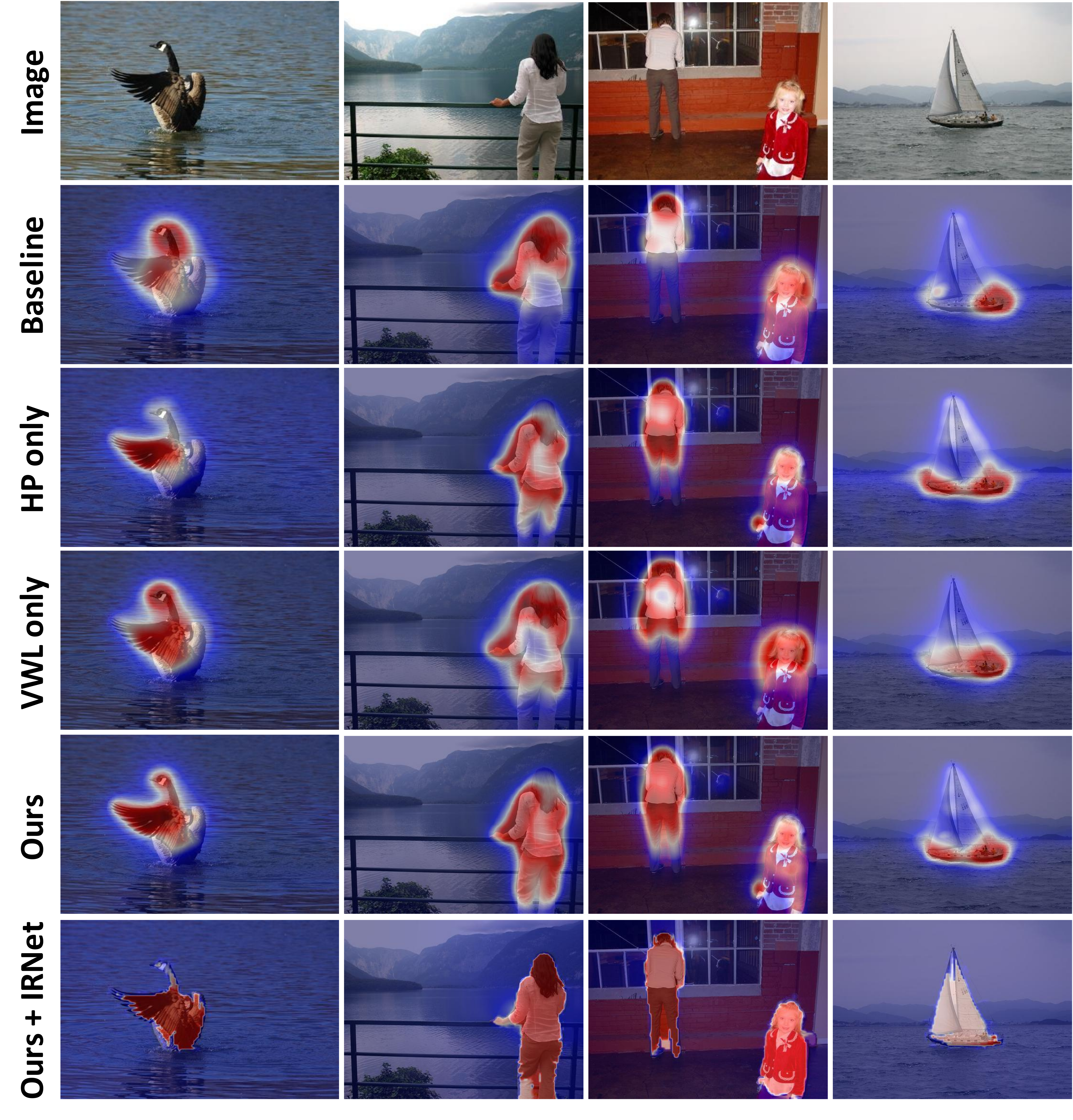}
  \caption{Visualization of the generated CAMs.}
  \label{fig_vis_cams_ablation}
\end{figure}

\paragraph{\textbf{Visual Ablation Results.}} Our intention of the proposed VWL and HP is to encourage the network to activate more object extents and fewer background regions, respectively. Though Table~\ref{tab_ablation} shows that the proposed methods could improve the quality of CAMs, we still want to explore their effects on the generated CAMs. Therefore, we further visualize the CAMs generated by baseline, baseline with only HP, baseline with only VWL and our method. The visualization results are presented in Fig.~\ref{fig_vis_cams_ablation}. It is observed that VWL typically discovers more object extents, while both of them tend to activate adjacent background around objects. HP could remarkably alleviate this drawback since it aggregates local discriminative information instead of the whole image. Our method, which combines VWL and HP, could jointly mine more object regions and reduce the unexpected background. Fig.~\ref{fig_vis_cams_ablation} also shows IRNet \citep{ahn2019weakly} could further dampen the falsely activated regions and diffuse the object regions, so the CAMs can align better with the semantic boundaries.

\begin{figure*}[htp]
  \centering
  \includegraphics[width=0.98\textwidth]{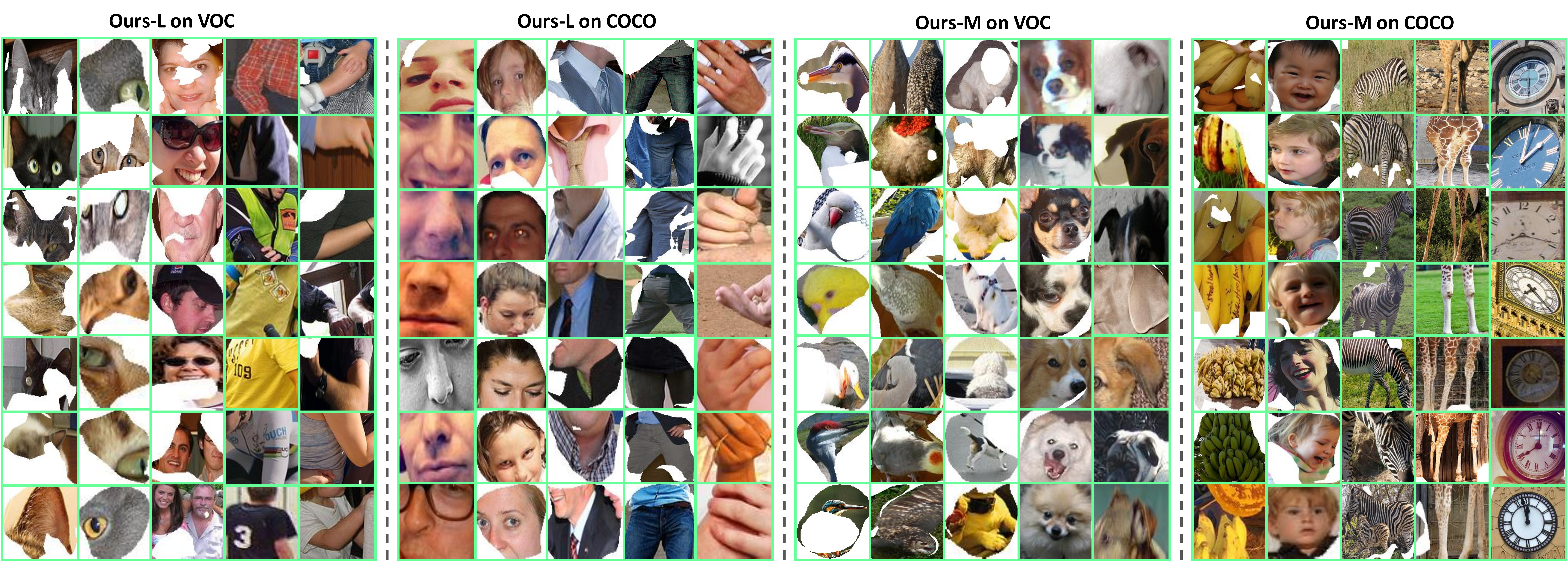}
  \caption{Visualization of the visual words learned by learning strategy and memory-bank strategy. Each column denotes the example images sampled from a visual word category.}
  \label{fig_vis_words}
\end{figure*}

\paragraph{\textbf{Codebook Analysis.}} To verify whether the learning-based and memory-bank strategy could learn a reasonable codebook, we visualize the learned visual words represented in the codebook by extracting their corresponding regions in an image. The visualization results are presented in Fig.~\ref{fig_vis_words}, where examples in each column are sampled from the images of a specific visual word. We show that both the learning-based strategy and memory-bank strategy could effectively learn visual word representations from images. For example, on the MS COCO dataset, our learning-based strategy could successfully decompose $\mathbf{person}$ to $\mathbf{face}$, $\mathbf{body}$, $\mathbf{hands}$ and $\mathbf{legs}$ \etc, which could be used to supervise the training of classification network and encourage more object extents to be discovered. Empirically, we also observe the learned visual words on MS COCO dataset typically consist of fewer noisy samples than the PASCAL VOC dataset, which indicates larger-scale dataset could benefit our visual words learning strategies.

\paragraph{\textbf{Codebook Initialization.}} \label{sec_codebook_init}
In this work, the codebook is simply randomly initialized. We do not use any extra pre-training or warm-up strategy for the codebook. To explore the impact of the initialization method, we present the performance of our method using {random sample} initialization (initializing the codebook with randomly sampled image features). The results are reported in Tab.~\ref{tab_init_2}. As shown in Tab.~\ref{tab_init_2}, the memory-bank strategy is not sensitive to the initialization method for the codebook, while the learning-based strategy achieves worse performance when using the random sample initialization. Technically, the codebook in the memory-bank strategy does not straightly affect the optimization process. Therefore, the impact of the initialization method is trivial. However, the codebook in the learning-based strategy is a trainable parameter and directly impacts the update process of the network parameters, thus notably affecting the performance of the generated CAM.

\begin{table}[h]

  \caption{The performance of the generated CAMs with different initialization methods. The results are evaluated on the PASCAL VOC $train$ set.}
  \label{tab_init_2}
  \centering
  \begin{tabular}{r|cc}
    \toprule
                                   & VWL-L & VWL-M \\ \midrule
    Random initialization          & 57.3  & 56.9  \\
    {Random sample} initialization & 55.8  & 56.6  \\ \bottomrule
  \end{tabular}
\end{table}

\paragraph{\textbf{Learning-based versus Memory-bank.}} The learning-based strategy (VWL-L) and the memory-bank strategy (VWL-M) are both inspired by the simple Bag of Visual Words model. Specifically, in VWL-L, the visual word representation for an input image is automatically learned with the devised loss functions, while VWL-M extracts the visual word representations by online clustering. In other words, VWL-L and VWL-M model an image implicitly and explicitly, respectively. Therefore, we empirically find VWL-M could yield better visual words than VWL-L. Besides, as discussed in Section~\ref{sec_codebook_init}, compared to VWL-M, VWL-L is slightly sensitive to the initialization of the codebook. However, on the efficiency side, due to the online reconstruction operation, the training process of VWL-M takes a slightly longer time than VWL-L. 

\begin{figure}[h]
  \centering
  \begin{subfigure}[b]{0.23\textwidth}
    \centering
    \includegraphics[height=0.88\textwidth]{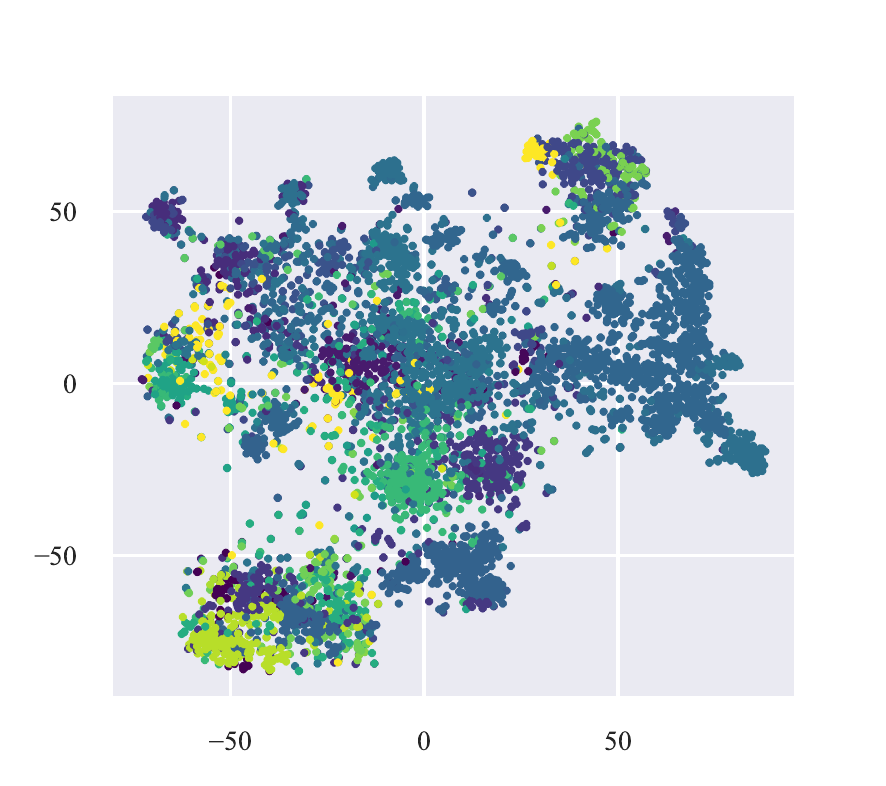}
    \caption{Learning-based strategy}
  \end{subfigure}
  \begin{subfigure}[b]{0.23\textwidth}
    \centering
    \includegraphics[height=0.88\textwidth]{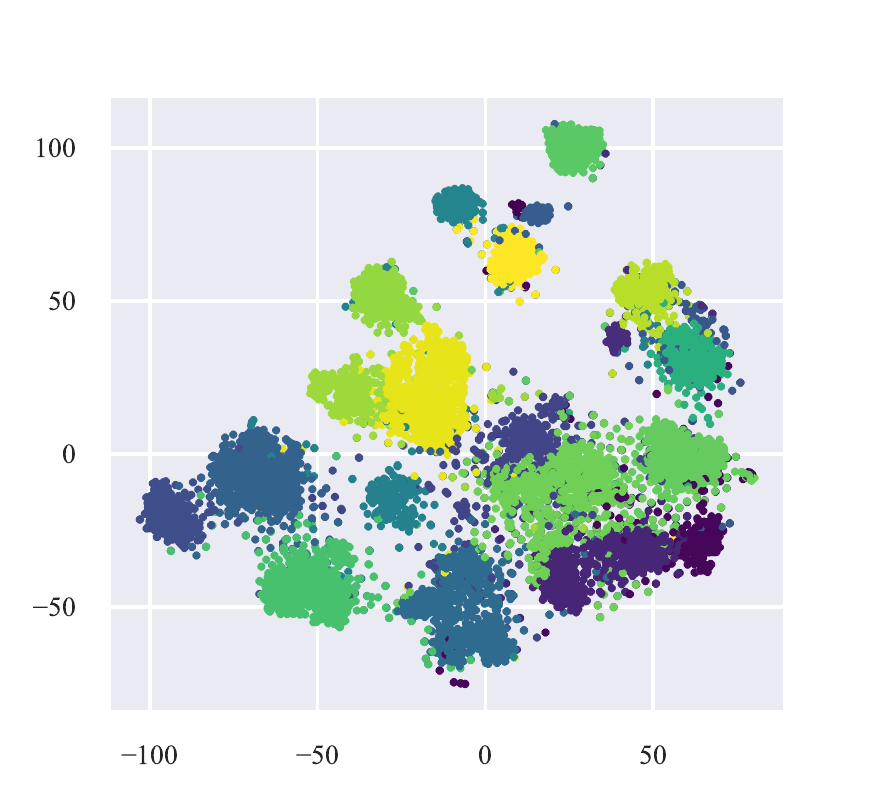}
    \caption{Memory-bank strategy}
  \end{subfigure}

  \caption{t-sne visualization of the generated visual words. Different colors denote the different visual words}
  \label{fig_tsne_1}
\end{figure}

\par To better understand the quality of the generated visual words by the learning-based and memory-bank strategy, in Fig.~\ref{fig_tsne_1}, we visualize the extracted visual word features with t-sne \citep{van2014accelerating}. The features for visualization are generated by averaging features of different visual word regions in each input image. As shown in Fig.~\ref{fig_tsne_1}, the clusters of VWL-L are more diverse than VWL-M's, \ie, VWL-M learns better visual words than VWL-L, which is attributed to the explicit modeling of visual words in VWL-M. We then use the extracted visual word frequencies of each image to predict the image-level labels. The classification accuracies are reported in Tab.~\ref{tab_word_freq_1}. VWL-M is still superior to VWL-L, demonstrating the better visual words learning capacity.

\begin{table}[h]

  \caption{The classification accuracies of the VWL-L and VWL-M. The performance is evaluated on the PASCAL VOC $train$ set.}
  \label{tab_word_freq_1}
  \centering
  \begin{tabular}{r|cc}
    \toprule
               & VWL-L & VWL-M \\ \midrule
    $Acc$ (\%) & 81.3  & 84.8  \\ \bottomrule
  \end{tabular}
\end{table}

\paragraph{\textbf{DeCov loss.}}
\begin{figure}[htp]
  \centering
  \includegraphics[width=0.22\textwidth]{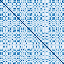}
  \includegraphics[width=0.22\textwidth]{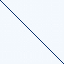}
  \caption{Visualization of the similarity matrix between each visual words \textbf{without} (\textit{left}) and \textbf{with} (\textit{right}) DeCov loss.}
  \label{fig_vis_codebook}
\end{figure}

\par In the loss function (Eq.~\eqref{eq_loss_aux}) for learning codebook in the learning-based strategy, we introduced the DeCov loss \citep{cogswell2015reducing} to reduce the redundancy of the learned visual word representations. In Table~\ref{tab_ablation}, we show that learning visual words with DeCov loss could improve the mIoU of generated CAMs on PASCAL VOC $train$ set from 55.1\% to 57.3\%. To further verify whether DeCov loss could eliminate the redundancy of the codebook, we visualized the similarity matrix of learned visual word representations. As presented in Fig.~\ref{fig_vis_codebook}, when using DeCov loss regularization, the cosine similarity between two different word representations is very close to 0. Taking the mIoU improvements in Table~\ref{tab_ablation} into consideration, we demonstrate our regularization loss in Eq.~\eqref{eq_loss_aux} could successfully reduce codebook redundancy and improve CAMs quality.

\begin{figure}[htp]

  \begin{minipage}{0.24\textwidth}
    \centering

    \begin{tabular}{l|cc}
      \toprule
                       & $train$ & $val$ \\ \midrule
      \textit{w/o} GAP & 41.1    & 39.9  \\
      \textit{w/} GAP  & 57.3    & 56.9  \\ \bottomrule
    \end{tabular}

  \end{minipage}
  \begin{minipage}{0.23\textwidth}
    \centering
    \includegraphics[width=0.95\textwidth]{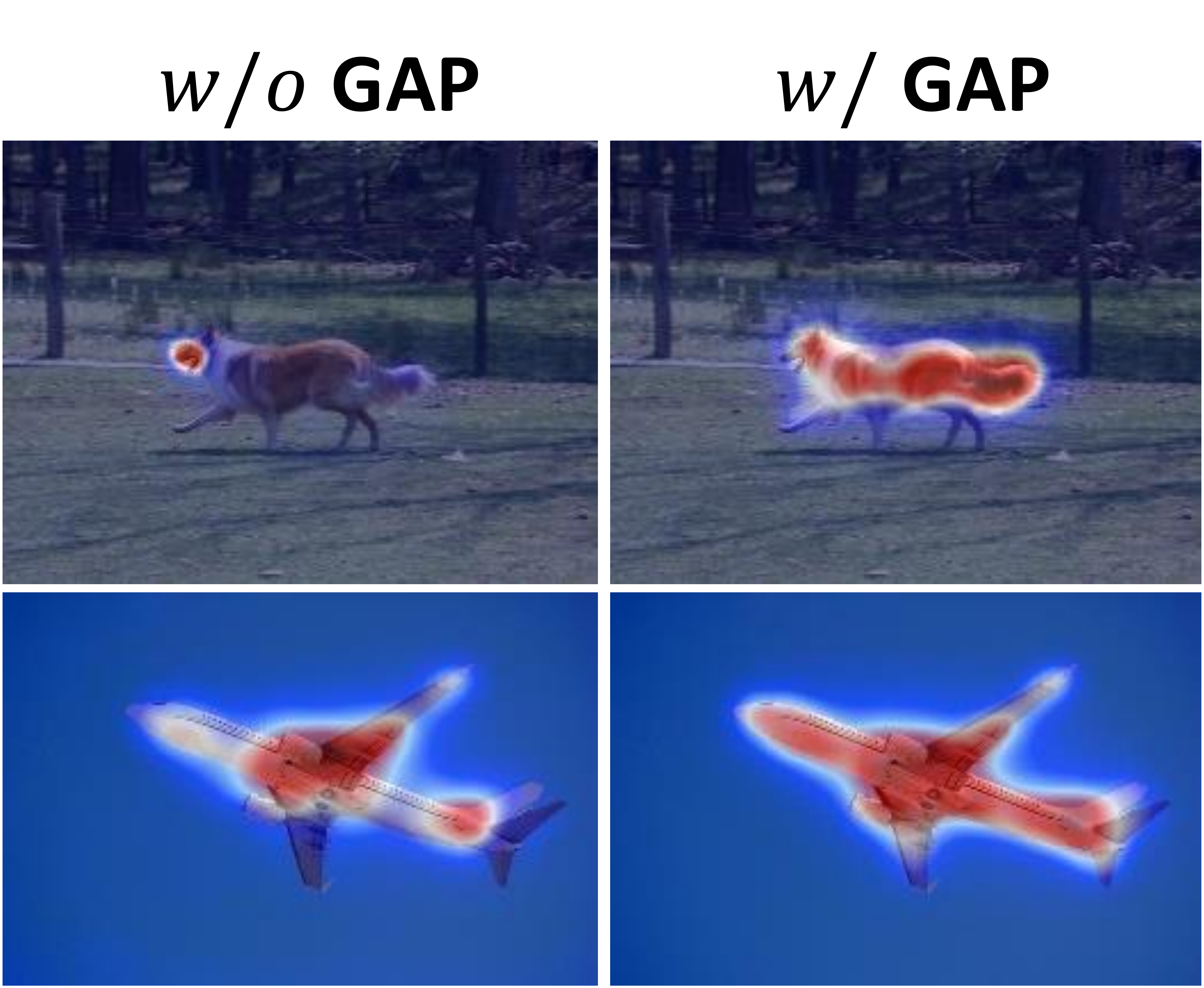}
  \end{minipage}

  \caption{Quantitative and visual results of hybrid pooling with (\textit{w/}) and without (\textit{w/o}) GAP.}
  \label{fig_gap_ablation}
\end{figure}

\paragraph{\textbf{GAP in HP for Object Completeness.}}
\par Some previous works \citep{kolesnikov2016seed, zhou2016learning} show that GAP tends to overestimate the object size while GMP tends to underestimate it. In the design of our hybrid pooling, we incorporate GAP to ensure object completeness. To verify the efficacy of GAP in HP, we present the quantitative and visual results of the generated CAMs using HP with and without GAP. As shown in Fig.~\ref{fig_gap_ablation}, using GAP in HP could bring a large mIoU improvement of about 16\%. The qualitative results also show that HP without GAP tends to discover only incomplete object regions while incorporating GAP could remarkably alleviate this problem.

\paragraph{\textbf{Parallel Branches in HP.}}
\par In the Hybrid Pooling module (HP), the parallel branches are used to produce multi-scale local information via max-pooling with different split sizes. Empirically, max-pooling with a small split size $r$ aggregates less foreground and background information, leading to more discriminative object activation in CAMs. On the contrary, max-pooling with a large $r$ aggregates more foreground and unexpected background information. In this work, we balance the background and foreground information aggregation by averaging the features with the different split sizes, \ie, parallel branches in HP. We present the impact of the split sizes in Tab.~\ref{tab_r}. Tab.~\ref{tab_r} shows that using max-pooling with the split size of $\{1,2,4\}$ in HP can activate the background and foreground regions in CAMs well, and achieve the best performance.

\begin{table}[h]
  \caption{Impact of the set of split size in HP. The results are evaluated on the PASCAL VOC $train$ set.}
  \label{tab_r}
  \centering
  \begin{tabular}{c|cccc}
    \toprule
    $r$     & $\{1\}$ & $\{1,2\}$ & $\{1,2,4\}$   & $\{1,2,4,8\}$ \\ \midrule
    $train$ & 56.2    & 56.1      & \textbf{57.3} & 57.0          \\ \bottomrule
  \end{tabular}
\end{table}

\paragraph{\textbf{Comparison to GWRP and LSE.}}
\par We compare the proposed HP with the global weighted rank pooling (GWRP) \citep{kolesnikov2016seed} and Log-Sum-Exp pooling (LSE) \citep{pinheiro2015image}. The experiments are conducted with ResNet101 \citep{he2016deep} as the backbone (without visual word learning). As shown in Tab.~\ref{tab_pooling}, our method clearly outperforms GWRP and LSE. Besides, compared to GWRP and LSE, our HP is easier to implement since it only incorporates avg-pooling and max-pooling.

\begin{table}[h]
  \caption{Comparison of the pooling method on the PASCAL VOC $train$ set.}
  \label{tab_pooling}
  \centering
  \begin{tabular}{c|cccc}
    \toprule
            & GAP  & GWRP & LSE  & Our HP        \\ \midrule
    $train$ & 49.5 & 50.6 & 51.6 & \textbf{54.0} \\ \bottomrule
  \end{tabular}
\end{table}

\paragraph{\textbf{GAP in Visual Word Learning.}}
As illustrated in Section \ref{sec_net_training}, we use GAP instead our HP for predicting the visual word labels. We conduct experiments using HP in visual word learning to explore its impact. As shown in Tab.~\ref{tab_vwl_pooling}, using HP in visual word learning (VWL-HP) also achieves notable improvements, demonstrating the effectiveness of VWL. Nevertheless, using GAP in visual word learning (VWL-GAP) could further outperform VWL-HP. We analyze the reason in Section~\ref{sec_net_training}: The pseudo visual words are generated based on all pixels of the feature maps. HP mainly considers partial discriminative information while GAP could aggregate all information. Therefore, we think GAP is the better pooling method for predicting visual words, which is also verified in Tab.~\ref{tab_vwl_pooling}.

\begin{table}[h]

  \caption{Comparison of the pooling method in the visual word learning process on the PASCAL VOC $train$ set.}
  \label{tab_vwl_pooling}
  \centering
  \begin{tabular}{c|ccc}
    \toprule
            & without VWL & VWL-HP & VWL-GAP       \\ \midrule
    $train$ & 54.0        & 55.5   & \textbf{57.3} \\ \bottomrule
  \end{tabular}
\end{table}

\subsection{Effect of Hyper-parameters}
\label{hyper_params}

\par This subsection presents the quantitative evaluation results of the generated CAMs on the PASCAL VOC $train$ and $val$ set with different hyper-parameter settings. All the results are evaluated and reported in mIoU.

\begin{table*}[htp]

  \caption{Impact of hyper-parameters.}
  \label{tab_params}
  \centering

  \begin{subtable}{0.27\textwidth}
    \setlength{\tabcolsep}{4pt}
    \begin{tabular}{l|llll}
      \toprule
              & \multicolumn{4}{c}{$k$}                                                                               \\ \cmidrule{2-5}
              & \multicolumn{1}{c}{128} & \multicolumn{1}{c}{256} & \multicolumn{1}{c}{384} & \multicolumn{1}{c}{512} \\ \midrule
      $train$ & 54.8                    & 57.3                    & 55.6                    & 55.4                    \\
      $val$   & 54.5                    & 56.9                    & 55.1                    & 54.9                    \\ \bottomrule
    \end{tabular}
    \caption{Number of visual words}
  \end{subtable}
  \begin{subtable}{0.21\textwidth}
    \setlength{\tabcolsep}{4pt}
    \begin{tabular}{llll}
      \toprule
      \multicolumn{4}{c}{ $\gamma$}                                                                 \\ \midrule
      \multicolumn{1}{c}{1} & \multicolumn{1}{c}{2} & \multicolumn{1}{c}{3} & \multicolumn{1}{c}{4} \\ \midrule
      55.2                  & 57.3                  & 56.6                  & 55.2                  \\
      54.5                  & 56.9                  & 55.8                  & 54.3                  \\ \bottomrule
    \end{tabular}
    \caption{Weight factor.}
  \end{subtable}
  \begin{subtable}{0.21\textwidth}
    \setlength{\tabcolsep}{4pt}
    \begin{tabular}{llll}
      \toprule
      \multicolumn{4}{c}{$\tau$}                                                                            \\ \midrule
      \multicolumn{1}{c}{0.6} & \multicolumn{1}{c}{0.8} & \multicolumn{1}{c}{1.0} & \multicolumn{1}{c}{1.2} \\ \midrule
      56.9                    & 57.1                    & 57.3                    & 56.7                    \\
      56.3                    & 56.4                    & 56.9                    & 55.9                    \\ \bottomrule
    \end{tabular}
    \caption{Temperature parameter.}
  \end{subtable}
  \begin{subtable}{0.22\textwidth}
    \setlength{\tabcolsep}{4pt}
    \begin{tabular}{llll}
      \toprule
      \multicolumn{4}{c}{$\rho$}                                                                                                    \\ \midrule
      \multicolumn{1}{c}{$1e^{-4}$} & \multicolumn{1}{c}{$1e^{-3}$} & \multicolumn{1}{c}{$1e^{-2}$} & \multicolumn{1}{c}{$1e^{-1}$} \\ \midrule
      55.6                          & 56.9                          & 56.3                          & 55.9                          \\
      55.0                          & 56.4                          & 55.6                          & 55.3                          \\ \bottomrule
    \end{tabular}
    \caption{Momentum coefficient.}
  \end{subtable}
\end{table*}

\paragraph{\textbf{Number of visual words $k$.}}
\par In Table~\ref{tab_params} (a), we present the impact of the number of visual words $k$ by setting it to $\{128,256,384,512\}$ and fixing other hyper-parameters (the classification network to generate CAMs is trained with the learning-based strategy). As observed in Table~\ref{tab_params} (a), our method with different $k$ could consistently outperform the baseline in Table~\ref{tab_ablation}, which demonstrates the effectiveness of our motivation. The best result is obtained with $k=256$.

\paragraph{\textbf{Weight factor $\gamma$.}}
\par The effect of the weight factor $\gamma$ in HP is presented in Table~\ref{tab_params} (b), which is used to trade off the GAP and GMP features. We observe that $\gamma=$ 2 works well, while the performance clearly decreases when $\gamma=1,4$ since the output feature will degrade to GMP or GAP when $\gamma$ is too small or big.

\paragraph{\textbf{Temperature parameter $\tau$.}}
\par In Eq.~\eqref{eq_p_ij}, we use a temperature parameter $\tau$ to control the smoothness of the visual word probabilities. As presented in Table~\ref{tab_params} (c), we empirically observe that $\tau=1.0,0.8$ are proper values for generating CAMs with higher quality.

\paragraph{\textbf{Momentum coefficient $\rho$.}}
\par For the memory-bank strategy, a momentum coefficient $\rho$ in Eq.~\eqref{eq_c_update} to manipulate the update rate of the codebook. In Table~\ref{tab_params} (d), we show that $\rho=1e^{-3}$ works finely since a large $\rho$ makes the codebook dependent on the features from the current batch, while a smaller $\rho$ means a slower update rate and may make the codebook not adaptable to training iterations.

\section{Conclusion}
\label{conclusion}
\par Previous CAMs typically only cover partial discriminative object regions and some unexpected background.
To tackle the first problem, we propose the visual words learning module. By enforcing the network to learn auxiliary visual words, more object regions could be activated. To perform unsupervised learning of visual words with only image-level labels, we devise the learning-based and memory-bank strategies to update the codebook.
To mitigate the second problem, we propose hybrid pooling, which aggregates local maximum and global average features to simultaneously reduce background regions in CAMs and ensure object completeness.
We experimentally demonstrated the superiority of our proposed method by surpassing recent state-of-the-art performance on the PASCAL VOC 2012 and MS COCO 2014 dataset.

\bibliographystyle{spbasic}      
\bibliography{vwl.bbl}   

\begin{thebibliography}{68}
\providecommand{\natexlab}[1]{#1}
\providecommand{\url}[1]{{#1}}
\providecommand{\urlprefix}{URL }
\expandafter\ifx\csname urlstyle\endcsname\relax
  \providecommand{\doi}[1]{DOI~\discretionary{}{}{}#1}\else
  \providecommand{\doi}{DOI~\discretionary{}{}{}\begingroup
  \urlstyle{rm}\Url}\fi
\providecommand{\eprint}[2][]{\url{#2}}

\bibitem[{Adams and Bischof(1994)}]{adams1994seeded}
Adams R, Bischof L (1994) Seeded region growing. IEEE Transactions on pattern
  analysis and machine intelligence 16(6):641--647

\bibitem[{Ahn and Kwak(2018)}]{ahn2018learning}
Ahn J, Kwak S (2018) Learning pixel-level semantic affinity with image-level
  supervision for weakly supervised semantic segmentation. In: Proceedings of
  the IEEE Conference on Computer Vision and Pattern Recognition, pp 4981--4990

\bibitem[{Ahn et~al.(2019)Ahn, Cho, and Kwak}]{ahn2019weakly}
Ahn J, Cho S, Kwak S (2019) Weakly supervised learning of instance segmentation
  with inter-pixel relations. In: Proceedings of the IEEE/CVF Conference on
  Computer Vision and Pattern Recognition, pp 2209--2218

\bibitem[{Arandjelovi{\'c} et~al.(2017)Arandjelovi{\'c}, Gronat, Torii, Pajdla,
  and Sivic}]{arandjelovic2017netvlad}
Arandjelovi{\'c} R, Gronat P, Torii A, Pajdla T, Sivic J (2017) Netvlad: Cnn
  architecture for weakly supervised place recognition. IEEE Transactions on
  Pattern Analysis and Machine Intelligence 40(6):1437--1451

\bibitem[{Araslanov and Roth(2020)}]{araslanov2020single}
Araslanov N, Roth S (2020) Single-stage semantic segmentation from image
  labels. In: Proceedings of the IEEE/CVF Conference on Computer Vision and
  Pattern Recognition, pp 4253--4262

\bibitem[{Badrinarayanan et~al.(2017)Badrinarayanan, Kendall, and
  Cipolla}]{badrinarayanan2017segnet}
Badrinarayanan V, Kendall A, Cipolla R (2017) Segnet: A deep convolutional
  encoder-decoder architecture for image segmentation. IEEE transactions on
  pattern analysis and machine intelligence 39(12):2481--2495

\bibitem[{Bearman et~al.(2016)Bearman, Russakovsky, Ferrari, and
  Fei-Fei}]{bearman2016s}
Bearman A, Russakovsky O, Ferrari V, Fei-Fei L (2016) What’s the point:
  Semantic segmentation with point supervision. In: European conference on
  computer vision, Springer, pp 549--565

\bibitem[{Chang et~al.(2020{\natexlab{a}})Chang, Wang, Hung, Piramuthu, Tsai,
  and Yang}]{chang2020mixup}
Chang YT, Wang Q, Hung WC, Piramuthu R, Tsai YH, Yang MH (2020{\natexlab{a}})
  Mixup-cam: Weakly-supervised semantic segmentation via uncertainty
  regularization. In: British Machine Vision Conference (BMVC)

\bibitem[{Chang et~al.(2020{\natexlab{b}})Chang, Wang, Hung, Piramuthu, Tsai,
  and Yang}]{chang2020weakly}
Chang YT, Wang Q, Hung WC, Piramuthu R, Tsai YH, Yang MH (2020{\natexlab{b}})
  Weakly-supervised semantic segmentation via sub-category exploration. In:
  Proceedings of the IEEE/CVF Conference on Computer Vision and Pattern
  Recognition, pp 8991--9000

\bibitem[{Chen et~al.(2015)Chen, Papandreou, Kokkinos, Murphy, and
  Yuille}]{chen2014semantic}
Chen LC, Papandreou G, Kokkinos I, Murphy K, Yuille AL (2015) Semantic image
  segmentation with deep convolutional nets and fully connected crfs. In:
  International Conference on Learning Representations

\bibitem[{Chen et~al.(2017)Chen, Papandreou, Kokkinos, Murphy, and
  Yuille}]{chen2017deeplab}
Chen LC, Papandreou G, Kokkinos I, Murphy K, Yuille AL (2017) Deeplab: Semantic
  image segmentation with deep convolutional nets, atrous convolution, and
  fully connected crfs. IEEE transactions on pattern analysis and machine
  intelligence 40(4):834--848

\bibitem[{Cogswell et~al.(2017)Cogswell, Ahmed, Girshick, Zitnick, and
  Batra}]{cogswell2015reducing}
Cogswell M, Ahmed F, Girshick R, Zitnick L, Batra D (2017) Reducing overfitting
  in deep networks by decorrelating representations. In: International
  Conference on Learning Representations

\bibitem[{Cordts et~al.(2016)Cordts, Omran, Ramos, Rehfeld, Enzweiler,
  Benenson, Franke, Roth, and Schiele}]{cordts2016cityscapes}
Cordts M, Omran M, Ramos S, Rehfeld T, Enzweiler M, Benenson R, Franke U, Roth
  S, Schiele B (2016) The cityscapes dataset for semantic urban scene
  understanding. In: Proceedings of the IEEE conference on computer vision and
  pattern recognition, pp 3213--3223

\bibitem[{Everingham et~al.(2010)Everingham, Van~Gool, Williams, Winn, and
  Zisserman}]{everingham2010pascal}
Everingham M, Van~Gool L, Williams CK, Winn J, Zisserman A (2010) The pascal
  visual object classes (voc) challenge. International journal of computer
  vision 88(2):303--338

\bibitem[{Fan et~al.(2020)Fan, Zhang, Tan, Song, and Xiao}]{fan2020cian}
Fan J, Zhang Z, Tan T, Song C, Xiao J (2020) Cian: Cross-image affinity net for
  weakly supervised semantic segmentation. In: Proceedings of the AAAI
  Conference on Artificial Intelligence, vol~34, pp 10762--10769

\bibitem[{Gao et~al.(2021)Gao, Cheng, Zhao, Zhang, Yang, and
  Torr}]{gao2021res2net}
Gao SH, Cheng MM, Zhao K, Zhang XY, Yang MH, Torr P (2021) Res2net: A new
  multi-scale backbone architecture. IEEE Transactions on Pattern Analysis and
  Machine Intelligence 43(2):652--662

\bibitem[{Gidaris et~al.(2020)Gidaris, Bursuc, Komodakis, P{\'e}rez, and
  Cord}]{gidaris2020learning}
Gidaris S, Bursuc A, Komodakis N, P{\'e}rez P, Cord M (2020) Learning
  representations by predicting bags of visual words. In: Proceedings of the
  IEEE/CVF Conference on Computer Vision and Pattern Recognition, pp 6928--6938

\bibitem[{Hariharan et~al.(2011)Hariharan, Arbel{\'a}ez, Bourdev, Maji, and
  Malik}]{hariharan2011semantic}
Hariharan B, Arbel{\'a}ez P, Bourdev L, Maji S, Malik J (2011) Semantic
  contours from inverse detectors. In: 2011 International Conference on
  Computer Vision, IEEE, pp 991--998

\bibitem[{He et~al.(2016)He, Zhang, Ren, and Sun}]{he2016deep}
He K, Zhang X, Ren S, Sun J (2016) Deep residual learning for image
  recognition. In: Proceedings of the IEEE conference on computer vision and
  pattern recognition, pp 770--778

\bibitem[{Hou et~al.(2017)Hou, Cheng, Hu, Borji, Tu, and Torr}]{hou2017deeply}
Hou Q, Cheng MM, Hu X, Borji A, Tu Z, Torr PH (2017) Deeply supervised salient
  object detection with short connections. In: Proceedings of the IEEE
  conference on computer vision and pattern recognition, pp 3203--3212

\bibitem[{Hou et~al.(2018)Hou, Jiang, Wei, and Cheng}]{hou2018self}
Hou Q, Jiang P, Wei Y, Cheng MM (2018) Self-erasing network for integral object
  attention. Advances in Neural Information Processing Systems 31:549--559

\bibitem[{Huang et~al.(2018)Huang, Wang, Wang, Liu, and Wang}]{huang2018weakly}
Huang Z, Wang X, Wang J, Liu W, Wang J (2018) Weakly-supervised semantic
  segmentation network with deep seeded region growing. In: Proceedings of the
  IEEE Conference on Computer Vision and Pattern Recognition, pp 7014--7023

\bibitem[{Jiang et~al.(2019)Jiang, Hou, Cao, Cheng, Wei, and
  Xiong}]{jiang2019integral}
Jiang PT, Hou Q, Cao Y, Cheng MM, Wei Y, Xiong HK (2019) Integral object mining
  via online attention accumulation. In: Proceedings of the IEEE/CVF
  International Conference on Computer Vision, pp 2070--2079

\bibitem[{Jo and Yu(2021)}]{jo2021puzzle}
Jo S, Yu IJ (2021) Puzzle-cam: Improved localization via matching partial and
  full features. In: 2021 IEEE International Conference on Image Processing
  (ICIP), pp 639--643

\bibitem[{Ke et~al.(2021)Ke, Hwang, and Yu}]{ke2021universal}
Ke TW, Hwang JJ, Yu SX (2021) Universal weakly supervised segmentation by
  pixel-to-segment contrastive learning. In: International Conference on
  Learning Representations

\bibitem[{Kim et~al.(2021)Kim, Han, and Kim}]{kim2021discriminative}
Kim B, Han S, Kim J (2021) Discriminative region suppression for
  weakly-supervised semantic segmentation. In: Proceedings of the AAAI
  Conference on Artificial Intelligence, vol~35, pp 1754--1761

\bibitem[{Kolesnikov and Lampert(2016)}]{kolesnikov2016seed}
Kolesnikov A, Lampert CH (2016) Seed, expand and constrain: Three principles
  for weakly-supervised image segmentation. In: European conference on computer
  vision, Springer, pp 695--711

\bibitem[{Kr{\"a}henb{\"u}hl and Koltun(2011)}]{krahenbuhl2011efficient}
Kr{\"a}henb{\"u}hl P, Koltun V (2011) Efficient inference in fully connected
  crfs with gaussian edge potentials. Advances in neural information processing
  systems 24:109--117

\bibitem[{Krizhevsky et~al.(2012)Krizhevsky, Sutskever, and
  Hinton}]{krizhevsky2012imagenet}
Krizhevsky A, Sutskever I, Hinton GE (2012) Imagenet classification with deep
  convolutional neural networks. Advances in neural information processing
  systems 25:1097--1105

\bibitem[{Lee et~al.(2021{\natexlab{a}})Lee, Kim, and Yoon}]{lee2021anti}
Lee J, Kim E, Yoon S (2021{\natexlab{a}}) Anti-adversarially manipulated
  attributions for weakly and semi-supervised semantic segmentation. In:
  Proceedings of the IEEE/CVF Conference on Computer Vision and Pattern
  Recognition, pp 4071--4080

\bibitem[{Lee et~al.(2021{\natexlab{b}})Lee, Yi, Shin, and Yoon}]{lee2021bbam}
Lee J, Yi J, Shin C, Yoon S (2021{\natexlab{b}}) Bbam: Bounding box attribution
  map for weakly supervised semantic and instance segmentation. In: Proceedings
  of the IEEE/CVF Conference on Computer Vision and Pattern Recognition, pp
  2643--2652

\bibitem[{Lee et~al.(2021{\natexlab{c}})Lee, Lee, Lee, and
  Shim}]{lee2021railroad}
Lee S, Lee M, Lee J, Shim H (2021{\natexlab{c}}) Railroad is not a train:
  Saliency as pseudo-pixel supervision for weakly supervised semantic
  segmentation. In: Proceedings of the IEEE/CVF Conference on Computer Vision
  and Pattern Recognition, pp 5495--5505

\bibitem[{Li et~al.(2021{\natexlab{a}})Li, Zhou, Li, Zhou, and
  Zhang}]{li2020group}
Li X, Zhou T, Li J, Zhou Y, Zhang Z (2021{\natexlab{a}}) Group-wise semantic
  mining for weakly supervised semantic segmentation. In: Proceedings of the
  AAAI Conference on Artificial Intelligence, vol~35, pp 1984--1992

\bibitem[{Li et~al.(2021{\natexlab{b}})Li, Kuang, Liu, Chen, and
  Zhang}]{li2021pseudo}
Li Y, Kuang Z, Liu L, Chen Y, Zhang W (2021{\natexlab{b}}) Pseudo-mask matters
  in weakly-supervised semantic segmentation. In: Proceedings of the IEEE/CVF
  International Conference on Computer Vision, pp 6964--6973

\bibitem[{Lin et~al.(2016)Lin, Dai, Jia, He, and Sun}]{lin2016scribblesup}
Lin D, Dai J, Jia J, He K, Sun J (2016) Scribblesup: Scribble-supervised
  convolutional networks for semantic segmentation. In: Proceedings of the IEEE
  conference on computer vision and pattern recognition, pp 3159--3167

\bibitem[{Lin et~al.(2019)Lin, Upchurch, and Bala}]{lin2020block}
Lin H, Upchurch P, Bala K (2019) Block annotation: Better image annotation with
  sub-image decomposition. In: Proceedings of the IEEE/CVF International
  Conference on Computer Vision (ICCV)

\bibitem[{Lin et~al.(2013)Lin, Chen, and Yan}]{lin2013network}
Lin M, Chen Q, Yan S (2013) Network in network. arXiv preprint arXiv:13124400

\bibitem[{Lin et~al.(2014)Lin, Maire, Belongie, Hays, Perona, Ramanan,
  Doll{\'a}r, and Zitnick}]{lin2014microsoft}
Lin TY, Maire M, Belongie S, Hays J, Perona P, Ramanan D, Doll{\'a}r P, Zitnick
  CL (2014) Microsoft coco: Common objects in context. In: European conference
  on computer vision, Springer, pp 740--755

\bibitem[{Liu et~al.(2019)Liu, Chen, Fieguth, Zhao, Chellappa, and
  Pietik{\"a}inen}]{liu2019bow}
Liu L, Chen J, Fieguth P, Zhao G, Chellappa R, Pietik{\"a}inen M (2019) From
  bow to cnn: Two decades of texture representation for texture classification.
  International Journal of Computer Vision 127(1):74--109

\bibitem[{Liu et~al.(2020)Liu, Wu, Wen, Shi, Qiu, and
  Cheng}]{liu2020leveraging}
Liu Y, Wu YH, Wen PS, Shi YJ, Qiu Y, Cheng MM (2020) Leveraging instance-,
  image-and dataset-level information for weakly supervised instance
  segmentation. IEEE Transactions on Pattern Analysis and Machine Intelligence

\bibitem[{Long et~al.(2015)Long, Shelhamer, and Darrell}]{long2015fully}
Long J, Shelhamer E, Darrell T (2015) Fully convolutional networks for semantic
  segmentation. In: Proceedings of the IEEE conference on computer vision and
  pattern recognition, pp 3431--3440

\bibitem[{Oh et~al.(2021)Oh, Kim, and Ham}]{oh2021background}
Oh Y, Kim B, Ham B (2021) Background-aware pooling and noise-aware loss for
  weakly-supervised semantic segmentation. In: Proceedings of the IEEE/CVF
  Conference on Computer Vision and Pattern Recognition, pp 6913--6922

\bibitem[{Papandreou et~al.(2015)Papandreou, Chen, Murphy, and
  Yuille}]{papandreou2015weakly}
Papandreou G, Chen LC, Murphy KP, Yuille AL (2015) Weakly-and semi-supervised
  learning of a deep convolutional network for semantic image segmentation. In:
  Proceedings of the IEEE international conference on computer vision, pp
  1742--1750

\bibitem[{Passalis and Tefas(2017)}]{passalis2017learning}
Passalis N, Tefas A (2017) Learning bag-of-features pooling for deep
  convolutional neural networks. In: 2017 IEEE International Conference on
  Computer Vision (ICCV), IEEE, pp 5766--5774

\bibitem[{Paszke et~al.(2019)Paszke, Gross, Massa, Lerer, Bradbury, Chanan,
  Killeen, Lin, Gimelshein, Antiga et~al.}]{paszke2019pytorch}
Paszke A, Gross S, Massa F, Lerer A, Bradbury J, Chanan G, Killeen T, Lin Z,
  Gimelshein N, Antiga L, et~al. (2019) Pytorch: An imperative style,
  high-performance deep learning library. Advances in Neural Information
  Processing Systems 32:8026--8037

\bibitem[{Pinheiro and Collobert(2015)}]{pinheiro2015image}
Pinheiro PO, Collobert R (2015) From image-level to pixel-level labeling with
  convolutional networks. In: Proceedings of the IEEE conference on computer
  vision and pattern recognition, pp 1713--1721

\bibitem[{Roy and Todorovic(2017)}]{roy2017combining}
Roy A, Todorovic S (2017) Combining bottom-up, top-down, and smoothness cues
  for weakly supervised image segmentation. In: Proceedings of the IEEE
  Conference on Computer Vision and Pattern Recognition, pp 3529--3538

\bibitem[{Ru et~al.(2021)Ru, Du, and Wu}]{ru2021learning}
Ru L, Du B, Wu C (2021) Learning visual words for weakly-supervised semantic
  segmentation. In: Proceedings of the Thirtieth International Joint Conference
  on Artificial Intelligence, {IJCAI-21}, pp 982--988

\bibitem[{Rubin(2019)}]{rubin2019essential}
Rubin DB (2019) Essential concepts of causal inference: a remarkable history
  and an intriguing future. Biostatistics \& Epidemiology 3(1):140--155

\bibitem[{Scarselli et~al.(2008)Scarselli, Gori, Tsoi, Hagenbuchner, and
  Monfardini}]{scarselli2008graph}
Scarselli F, Gori M, Tsoi AC, Hagenbuchner M, Monfardini G (2008) The graph
  neural network model. IEEE transactions on neural networks 20(1):61--80

\bibitem[{Sculley(2010)}]{sculley2010web}
Sculley D (2010) Web-scale k-means clustering. In: Proceedings of the 19th
  international conference on World wide web, pp 1177--1178

\bibitem[{Song et~al.(2019)Song, Huang, Ouyang, and Wang}]{song2019box}
Song C, Huang Y, Ouyang W, Wang L (2019) Box-driven class-wise region masking
  and filling rate guided loss for weakly supervised semantic segmentation. In:
  Proceedings of the IEEE/CVF Conference on Computer Vision and Pattern
  Recognition, pp 3136--3145

\bibitem[{Sun et~al.(2020)Sun, Wang, Dai, and Van~Gool}]{sun2020mining}
Sun G, Wang W, Dai J, Van~Gool L (2020) Mining cross-image semantics for weakly
  supervised semantic segmentation. In: European Conference on Computer Vision,
  Springer, pp 347--365

\bibitem[{Van Der~Maaten(2014)}]{van2014accelerating}
Van Der~Maaten L (2014) Accelerating t-sne using tree-based algorithms. The
  Journal of Machine Learning Research 15(1):3221--3245

\bibitem[{Vernaza and Chandraker(2017)}]{vernaza2017learning}
Vernaza P, Chandraker M (2017) Learning random-walk label propagation for
  weakly-supervised semantic segmentation. In: Proceedings of the IEEE
  conference on computer vision and pattern recognition, pp 7158--7166

\bibitem[{Wang et~al.(2020{\natexlab{a}})Wang, Liu, Ma, and
  Yang}]{wang2020weakly}
Wang X, Liu S, Ma H, Yang MH (2020{\natexlab{a}}) Weakly-supervised semantic
  segmentation by iterative affinity learning. International Journal of
  Computer Vision 128(6):1736--1749

\bibitem[{Wang et~al.(2020{\natexlab{b}})Wang, Zhang, Kan, Shan, and
  Chen}]{wang2020self}
Wang Y, Zhang J, Kan M, Shan S, Chen X (2020{\natexlab{b}}) Self-supervised
  equivariant attention mechanism for weakly supervised semantic segmentation.
  In: Proceedings of the IEEE/CVF Conference on Computer Vision and Pattern
  Recognition, pp 12275--12284

\bibitem[{Wei et~al.(2017)Wei, Feng, Liang, Cheng, Zhao, and
  Yan}]{wei2017object}
Wei Y, Feng J, Liang X, Cheng MM, Zhao Y, Yan S (2017) Object region mining
  with adversarial erasing: A simple classification to semantic segmentation
  approach. In: Proceedings of the IEEE conference on computer vision and
  pattern recognition, pp 1568--1576

\bibitem[{Wu et~al.(2021)Wu, Huang, Gao, Wei, Wei, Luo, and
  Liu}]{wu2021embedded}
Wu T, Huang J, Gao G, Wei X, Wei X, Luo X, Liu CH (2021) Embedded
  discriminative attention mechanism for weakly supervised semantic
  segmentation. In: Proceedings of the IEEE/CVF Conference on Computer Vision
  and Pattern Recognition, pp 16765--16774

\bibitem[{Wu et~al.(2018)Wu, Xiong, Yu, and Lin}]{wu2018unsupervised}
Wu Z, Xiong Y, Yu SX, Lin D (2018) Unsupervised feature learning via
  non-parametric instance discrimination. In: Proceedings of the IEEE
  Conference on Computer Vision and Pattern Recognition, pp 3733--3742

\bibitem[{Xu et~al.(2021)Xu, Ouyang, Bennamoun, Boussaid, Sohel, and
  Xu}]{xu2021leveraging}
Xu L, Ouyang W, Bennamoun M, Boussaid F, Sohel F, Xu D (2021) Leveraging
  auxiliary tasks with affinity learning for weakly supervised semantic
  segmentation. In: Proceedings of the IEEE/CVF International Conference on
  Computer Vision, pp 6984--6993

\bibitem[{Yao et~al.(2021)Yao, Chen, Xie, Zhang, Shen, Wu, Tang, and
  Zhang}]{yao2021non}
Yao Y, Chen T, Xie GS, Zhang C, Shen F, Wu Q, Tang Z, Zhang J (2021)
  Non-salient region object mining for weakly supervised semantic segmentation.
  In: Proceedings of the IEEE/CVF Conference on Computer Vision and Pattern
  Recognition, pp 2623--2632

\bibitem[{Zhang et~al.(2020{\natexlab{a}})Zhang, Xiao, Wei, Sun, and
  Huang}]{zhang2020reliability}
Zhang B, Xiao J, Wei Y, Sun M, Huang K (2020{\natexlab{a}}) Reliability does
  matter: An end-to-end weakly supervised semantic segmentation approach. In:
  Proceedings of the AAAI Conference on Artificial Intelligence, vol~34, pp
  12765--12772

\bibitem[{Zhang et~al.(2020{\natexlab{b}})Zhang, Zhang, Tang, Hua, and
  Sun}]{zhang2020causal}
Zhang D, Zhang H, Tang J, Hua XS, Sun Q (2020{\natexlab{b}}) Causal
  intervention for weakly-supervised semantic segmentation. Advances in Neural
  Information Processing Systems 33

\bibitem[{Zhang et~al.(2018)Zhang, Wei, Feng, Yang, and
  Huang}]{zhang2018adversarial}
Zhang X, Wei Y, Feng J, Yang Y, Huang TS (2018) Adversarial complementary
  learning for weakly supervised object localization. In: Proceedings of the
  IEEE Conference on Computer Vision and Pattern Recognition, pp 1325--1334

\bibitem[{Zheng et~al.(2015)Zheng, Jayasumana, Romera-Paredes, Vineet, Su, Du,
  Huang, and Torr}]{zheng2015conditional}
Zheng S, Jayasumana S, Romera-Paredes B, Vineet V, Su Z, Du D, Huang C, Torr PH
  (2015) Conditional random fields as recurrent neural networks. In:
  Proceedings of the IEEE international conference on computer vision, pp
  1529--1537

\bibitem[{Zhou et~al.(2016)Zhou, Khosla, Lapedriza, Oliva, and
  Torralba}]{zhou2016learning}
Zhou B, Khosla A, Lapedriza A, Oliva A, Torralba A (2016) Learning deep
  features for discriminative localization. In: Proceedings of the IEEE
  conference on computer vision and pattern recognition, pp 2921--2929

\bibitem[{Zhuang et~al.(2019)Zhuang, Zhai, and Yamins}]{zhuang2019local}
Zhuang C, Zhai AL, Yamins D (2019) Local aggregation for unsupervised learning
  of visual embeddings. In: Proceedings of the IEEE/CVF International
  Conference on Computer Vision, pp 6002--6012

\end{thebibliography}


\end{document}